%% file: main.tex
\documentclass{article}

\usepackage[preprint]{neurips_2025}

\usepackage[utf8]{inputenc} % allow utf-8 input
\usepackage[T1]{fontenc}    % use 8-bit T1 fonts
\usepackage{hyperref}       % hyperlinks
\usepackage{url}            % simple URL typesetting
\usepackage{booktabs}       % professional-quality tables
\usepackage{amsfonts}       % blackboard math symbols
\usepackage{nicefrac}       % compact symbols for 1/2, etc.
\usepackage{microtype}      % microtypography
\usepackage{xcolor}         % colors

\usepackage{graphicx}
\usepackage{colortbl}
\usepackage{multirow}
\usepackage{multicol}
\usepackage{bm}
\usepackage{subcaption}
\usepackage{diagbox}
\usepackage{amsmath,mathtools}
\usepackage{amssymb}
\usepackage{animate}
\usepackage{setspace}
\usepackage{booktabs} 
\usepackage{tcolorbox}

\usepackage[inline]{enumitem}
\usepackage{tabularx}
\usepackage{xspace}
\usepackage{interval}
\usepackage{siunitx}
\usepackage{epigraph}
\usepackage{inconsolata}
\usepackage{caption}
\usepackage{float}
\usepackage{soul}
\usepackage{pifont}
\usepackage{makecell} 
\usepackage{wrapfig}

\newcommand{\system}{\textit{UniGen}\xspace}
\newcommand{\systempt}{\textit{UniGen-PT}\xspace}
\newcommand{\systemsft}{\textit{UniGen-SFT}\xspace}
\newcommand{\systemdpo}{\textit{UniGen-DPO}\xspace}
\newcommand{\cotfull}{\textit{Chain-of-Thought Verification}\xspace}
\newcommand{\cotshort}{\textit{CoT-V}\xspace}
\newcommand{\ra}[1]{\renewcommand{\arraystretch}{#1}}
\newcommand*{\eg}{\emph{e.g.}\@\xspace}

\newcommand*{\ie}{\emph{i.e.}\@\xspace}

\usepackage[capitalize]{cleveref}
\crefname{section}{Sec.}{Secs.}
\Crefname{section}{Section}{Sections}
\Crefname{table}{Table}{Tables}
\crefname{table}{Tab.}{Tabs.}
\crefname{prompt}{Prompt.}{Prompts.}

\definecolor{lightgrey}{RGB}{211, 211, 211}
\definecolor{risecolor}{RGB}{0, 139, 139}
\definecolor{dropcolor}{RGB}{119, 136, 153}
\definecolor{lightblue}{rgb}{0.93, 0.96, 1.0}

\newcounter{prompt}
\renewcommand{\theprompt}{\arabic{prompt}}
\newcommand{\prompt}[3]{
\refstepcounter{prompt}
\begin{tcolorbox}[colback=lightblue!35, colframe=white!45!black, title={ Prompt.~\theprompt:~#1}]
#2
\label{#3}
\end{tcolorbox}
}

\title{\system: Enhanced Training\,\&\,Test-Time Strategies for \\ Unified Multimodal Understanding and Generation}
\author{Rui Tian$^{1\,2\,\circ}$, \,Mingfei Gao$^{1\,\circ}$, \,Mingze Xu$^{1\,\circ}$, \\ \,\textbf{Jiaming Hu$^1$, \,Jiasen Lu$^1$, \,Zuxuan Wu$^{2\,\dagger}$, \,Yinfei Yang$^1$, \,Afshin Dehghan$^{1\,\dagger}$} \\
$^1$Apple \quad $^2$Fudan University  \\
\small{\texttt{\{mgao22,mingze\_xu2,adehghan\}@apple.com}, \, \texttt{\{rtian23,zxwu\}@fudan.edu.cn}}, \\
$^\circ$First authors; \,$^\dagger$Corresponding authors
}

\begin{document}

\maketitle

\input{sec/0_abstract}
\input{sec/1_intro}

\input{sec/2_related}
\input{sec/3_methods}
\input{sec/4_experiments}
\input{sec/6_conclusion}

\appendix
\clearpage
\input{sec/X_supple}

%%%%%%%%%%%%%%%%%%%%%%%%%%%%%%%%%%%%%%%%%%%%%%%%%%%%%%%%%%%%

\bibliographystyle{plain}
\bibliography{main}

\end{document}

%% file: sec/0_abstract.tex
\begin{abstract}
We introduce \textbf{\system}, a unified multimodal large language model (MLLM) capable of image understanding and generation. We study the full training pipeline of \system from a data-centric perspective, including multi-stage pre-training, supervised fine-tuning, and direct preference optimization. More importantly, we propose a new \textbf{\cotfull (\cotshort)} strategy for test-time scaling, which significantly boosts \system's image generation quality using a simple \textit{Best-of-N} test-time strategy. Specifically, \cotshort enables \system to act as both image generator and verifier at test time, assessing the semantic alignment between a text prompt and its generated image in a step-by-step CoT manner. Trained entirely on open-source datasets across all stages, \system achieves state-of-the-art performance on a range of image understanding and generation benchmarks, with a final score of $0.78$ on \textsc{GenEval} and $85.19$ on \textsc{DPG-Bench}. Through extensive ablation studies, our work provides actionable insights and addresses key challenges in the full life cycle of building unified MLLMs, contributing meaningful directions to the future research.
\end{abstract}

\begin{figure}[H]
    \centering
    \vspace{-5pt}
    \includegraphics[width=0.92\linewidth]{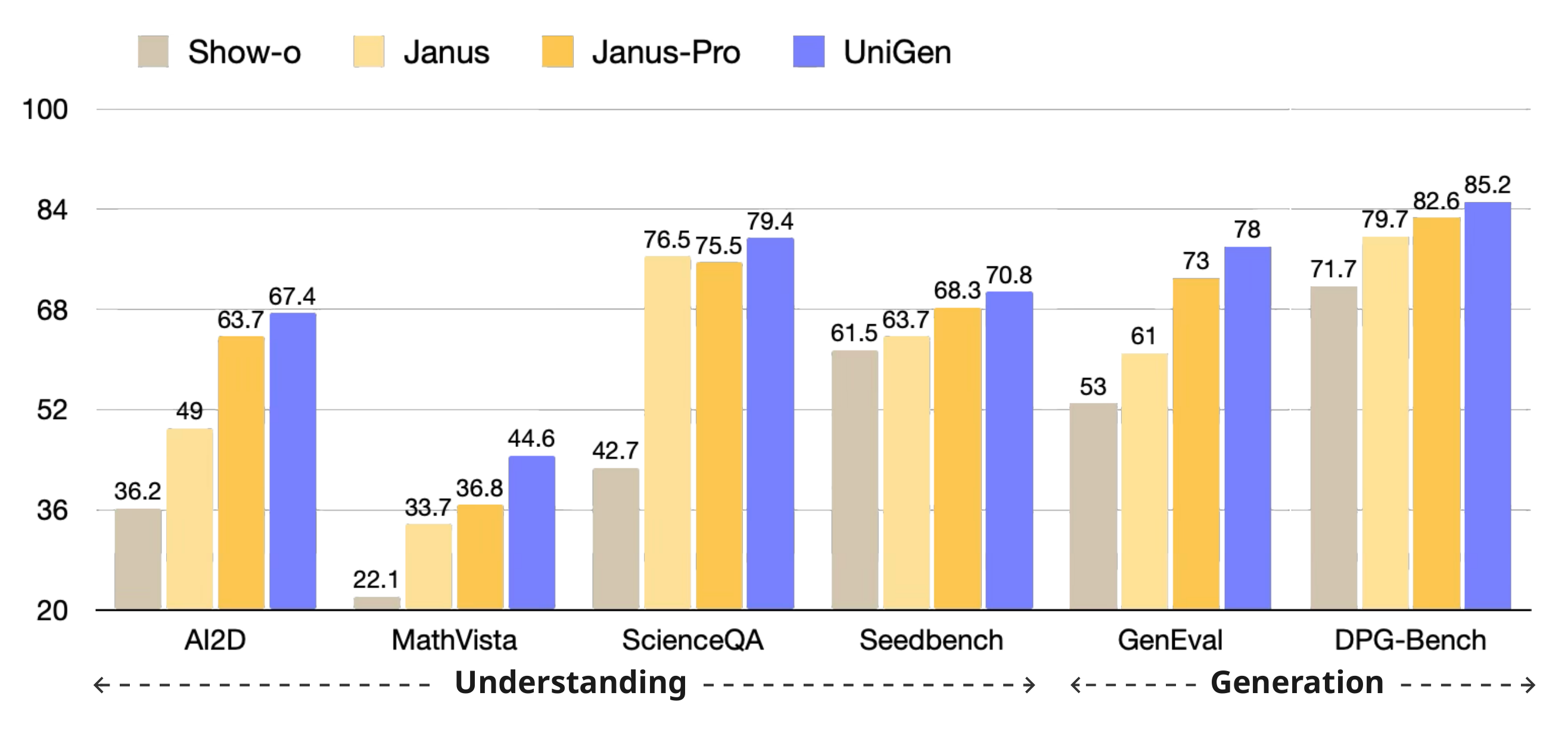}
    \vspace{-10pt}
    \caption{
        \textbf{Comparison against state-of-the-art unified MLLMs.} \textit{\system-1.5B} outperforms Show-o-1.3B, Janus-1.3B and Janus-Pro-1.5B across understanding and generation benchmarks.
    }
    \vspace{-5pt}
    \label{fig:benchmarks}
\end{figure}

%% file: sec/1_intro.tex
\vspace{-2pt}
\section{Introduction}
\label{sec:introduction}
\vspace{-5pt}

Unifying understanding and generation within a single framework represents a key step toward general-purpose artificial intelligence models~\cite{gpt4o}.
Pioneering work~\cite{team2024chameleon,ge2024seed,ye2024x,xie2024show,wu2024janus,chen2025janus} has made encouraging progress but relies on distinct training recipes and internal datasets.
More importantly, they have yet to demonstrate good practice in wisely collaborating these two capabilities within a unified architecture to achieve substantial performance gains.
We advance the development of unified multimodal large language models (MLLMs)
by carefully studying the impact of their training recipes across different stages and proposing optimizations to improve
both image understanding and generation.
We further explore leveraging test-time interaction between understanding and generation tasks, selecting images with higher quality by using our unified MLLM as the self-verifier.

Specifically, we introduce \textbf{\system}, a unified MLLM for image understanding and generation.
To shed light on the impact of different training stages, we walk through the entire life cycle of the model development, including multi-stage pretraining, supervised fine-tuning, and direct preference optimization~\cite{rafailov2023direct,wang2024enhancing}.
We ablate the impact of each training stage and their design choices from a data-centric perspective,
and draw insightful lessons for building advanced unified MLLMs.
Unlike state-of-the-art models~\cite{xie2024show,chen2025janus} that rely on large-scale internal datasets, we curate new data mixtures across training stages by using only open-source images.
We show that models trained on publicly available data can also achieve competitive results.

To further enhance image generation quality,
we propose a new \textbf{\cotfull (\cotshort)} strategy for test-time scaling.
The key idea is to leverage \system's inherent understanding ability as a self-verifier to assess the quality of its own generated images.
Specifically, during inference, \system produces $N$ images for a given text prompt,
while \cotshort progressively evaluates semantic coherence between each image-text pair and selects the best.
With only lightweight fine-tuning (\eg, 500 training steps),
\system is able to achieve the reasoning capability,
thinking step-by-step to verify each atomic fact according to the prompt and each generated image.
Importantly, this CoT verification seamlessly enhances \system's image generation quality while preserving its general understanding performance.
In this way, we collaborate the understanding and generation capabilities within a unified MLLM, substantially boosting the text-to-image generation quality using a simple \emph{Best-of-N} strategy~\cite{wang2025visualprm,zhu2025internvl3} and self-verification~\cite{weng2023large,chen2025sets,huang2025efficient}.
Our experiments show that \system's performance is consistently improved across various image generation benchmarks.

We evaluate \system on various understanding and generation tasks, as shown in Fig.~\ref{fig:benchmarks}. For image understanding, \system outperforms comparable unified MLLMs (\eg Show-o~\cite{xie2024show} and Janus-Pro~\cite{chen2025janus}) across benchmarks and even ties with some strong understanding specialist models, such as LLaVA-OV \cite{li2024llava} and MM1.5 \cite{zhang2024mm1}, as displayed in Table~\ref{tab:mmu_cmp}.
For text-to-image generation, \system obtains $0.78$ on \textsc{GenEval} and $85.19$ on \textsc{DPG-Bench} using only open-source data, surpassing state-of-the-art unified MLLMs~\cite{xie2024show,wu2024janus,chen2025janus} by a clear margin.

%% file: sec/2_related.tex
\vspace{-5pt}
\section{Related Work}
\label{sec:relate_work}
\vspace{-7pt}

\noindent \textbf{Multimodal Large Language Models (MLLMs)} have advanced significantly in image~\cite{liu2023visual,abdin2024phi,deitke2024molmo,liu2024nvila,wang2024qwen2,zhu2025internvl3} and video understanding~\cite{xu2024slowfast,llava178k,liu2024oryx,zohar2024apollo,xu2025slowfast,zhang2025videollama3}.
Their architecture typically consists of a vision encoder~\cite{radford2021learning,zhai2023sigmoid,tschannen2025siglip} to extract visual features, a projector~\cite{li2023blip,alayrac2022flamingo} to align image-text embeddings,
and a large language model (LLM)~\cite{abdin2024phi,touvron2023llama,qwen2.5,deepseekai2024deepseekv3technicalreport} to generate responses.
Early work focuses on pre-training using large-scale vision-language corpus~\citep{mckinzie2024mm1,tong2024cambrian}, then moves to carefully curated instructional datasets for supervised fine-tuning~\citep{li2024llava,zhang2024mm1} and reinforcement learning~\cite{ivison2024unpacking,rafailov2023direct}.
Recently, enabling MLLMs to output explicit reasoning trajectories has become a promising research direction~\cite{openai-o1,guo2025deepseek,team2025kimi}. They explore strategies, such as chain-of-thought (CoT) prompting~\cite{dong2024insight,yao2024mulberry}, reinforcement learning~\cite{2023llavarlhf,xiong2024llavacritic}, and test-time scaling~\cite{zhu2025internvl3,wang2025visualprm} to enhance the visual reasoning capabilities of MLLMs.

\noindent \textbf{Unified Understanding and Generation} aims to combine visual understanding and generation within a single MLLM framework~\cite{4m,team2024chameleon,qu2024tokenflow,wang2025mint,lu2022unified,lu2023uio2}. This is often achieved by jointly optimizing LLMs with multimodal objectives and generation-specific losses, such as autoregressive decoding~\cite{wu2024janus}, diffusion~\cite{zhou2024transfusion}, flow-matching~\cite{ma2024janusflow}, and masked image prediction~\cite{yang2025hermesflow,xie2024show}.
Visual tokenizers~\cite{van2017neural,esser2021taming,yu2023language,mentzer2023finite,zhao2024image} are critical for enabling both semantic understanding and high-fidelity generation. Recent efforts explore both decoupled encoders~\cite{tong2024metamorph,wu2024janus} and unified tokenizers~\cite{wu2024vila,jiao2025unitoken,qu2024tokenflow} for better task balancing.
Integrating CoT into visual generation emerges as a promising strategy.
PARM~\cite{guo2025can} scales test-time computation by introducing a new verification process.
MINT~\cite{wang2025mint}, ImageGen-CoT~\cite{liao2025imagegen}, and Got~\cite{fang2025got} leverage multimodal reasoning to perform prompt planning, generation, reflection, and refinement.
However, using CoT for unified understanding and generation remains underexplored. In this work, \system adopts a CoT-based self-verification strategy via Best-of-N selection during test-time scaling, which significantly improves the image generation performance. 

%% file: sec/3_methods.tex
\vspace{-5pt}
\section{Recipe for Building \system}
\label{sec:unigen}
\vspace{-5pt}

\begin{wrapfigure}{r}{0.55\linewidth}
    \centering
    \vspace{-20pt}
    \includegraphics[width=\linewidth]{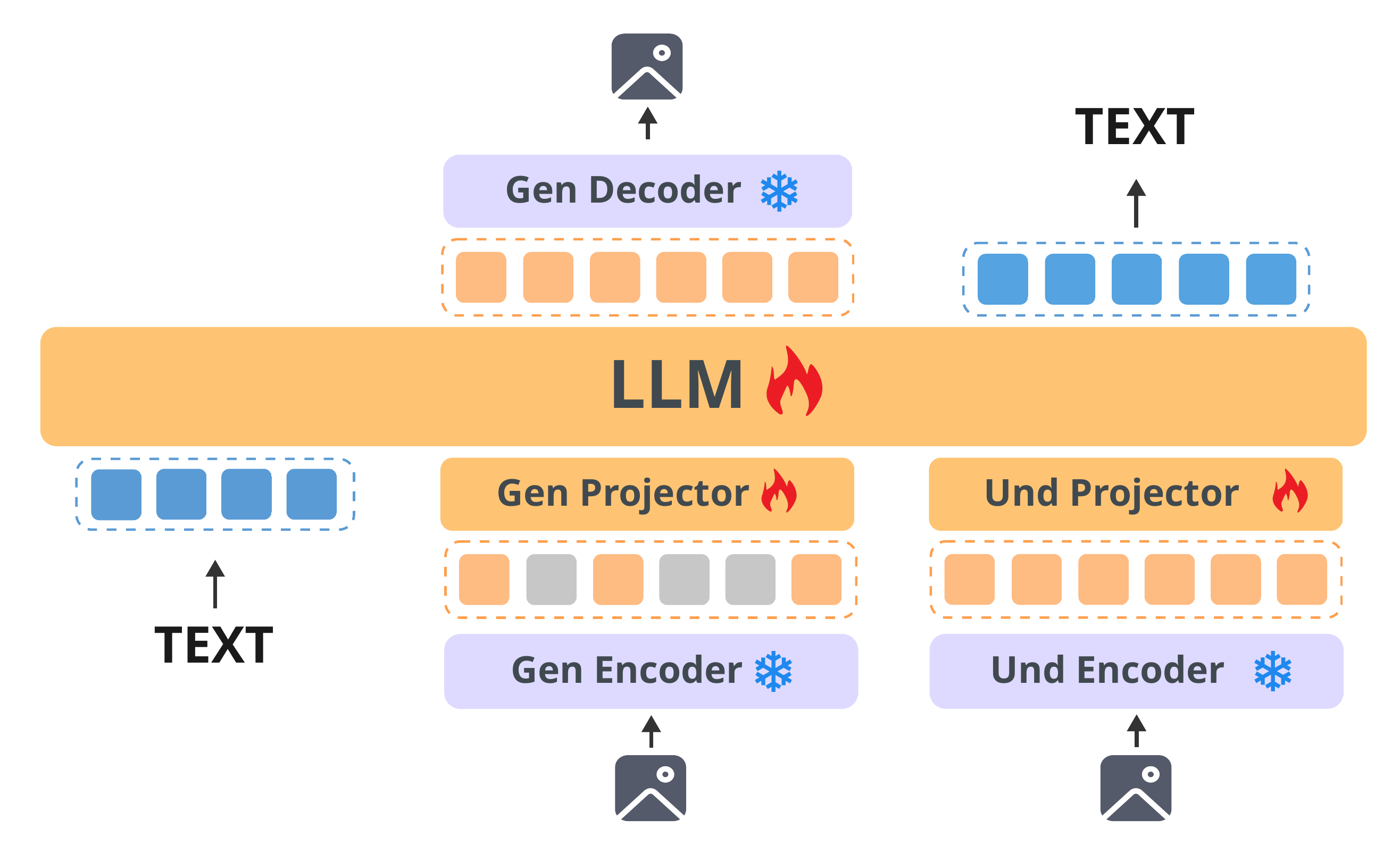}
    \caption{
        \textbf{The architecture of \system}, which is based on an autoregressive LLM and decoupled vision encoders for image understanding and generation tasks.
    }
    \vspace{-10pt}
    \label{fig:unigen_arch}
\end{wrapfigure}

\subsection{Architecture}
\label{sec:unigen_arch}
\vspace{-5pt}

As shown in Fig.~\ref{fig:unigen_arch}, we unify the image understanding and generation tasks into a pre-trained LLM, while decoupling their visual encoding into continuous and discrete embedding spaces.
This design is inspired by prior work~\cite{wu2024janus}, showing that continuous embeddings effectively preserve spatial details for understanding tasks, whereas discrete tokens are inherently well-suited for image generation.

\textbf{For image understanding}, we follow the LLaVA~\cite{liu2023visual} workflow and adopt the next-token prediction paradigm.
Given an input image $X^{U}$, the understanding encoder $\mathbf{Enc}^{U}$ (\eg, SigLIP~\cite{zhai2023sigmoid}) extracts its feature as a vector of continuous tokens $\mathcal{X}^{U}=\mathbf{Enc}^{U}(X^{U})$.
The projector $\mathbf{P}^{U}$ aligns the image and text embeddings into the same space,
then the embeddings are fed into LLM as inputs.
We compute the understanding loss using the vanilla autoregressive training objective $\mathcal{L}_{und}$.
To preserve the LLM's language modeling capability, we also train \system with text-only data
and backpropagate the corresponding loss $\mathcal{L}_{text}$.

\textbf{For text-to-image generation},
we employ the masked token prediction~\cite{chang2022maskgit} as our training objectives.
Unlike the autoregressive decoding for text tokens,
this paradigm enables models to generate multiple image tokens in parallel,
significantly accelerating the generation process.
\textit{During training}, for each image $X^{G} \in \mathbb{R}^{H\times W}$,
the generation encoder $\mathbf{Enc}^{G}$ (\eg, MAGVIT-v2~\cite{yulanguage}) tokenizes it into a sequence of discrete tokens $\mathcal{X}^{G}$ of length $N = {H}/{d_s}\cdot{W}/{d_s}$,
where $d_s$ refers to the spatial down-sampling factor of $\mathbf{Enc}^{G}$.
Then, given a masking ratio $\eta$ according to the scheduling function $\gamma(\cdot)$, we randomly sample a binary mask $\mathcal{M}(\eta) = [m_0, \cdots, m_{N-1}]$, where $\eta * N$ positions are uniformly set to $1$ and others are set to $0$.
For each position $i$ where $m_i$ equals to $1$, we replace
its corresponding discrete image token $\mathcal{X}^{G}_i$ with a special mask token $\mathtt{[MASK]}$  to form the final input image sequence. 
Finally, we prepend the textual tokens (\eg, image classes or captions) with the masked sequence $\mathcal{X}^{\mathcal{M}}$.
% The training objective $\mathcal{L}_{gen}$ for visual generation can be denoted as
% %
% \begin{equation}
% \small
% \mathcal{L}_{gen} = -\mathbb{E}_{\mathcal{X}^{\mathcal{M}} \in \mathcal{D}}[\sum_{i=0}^N m_i \cdot \log p(y_i | \mathcal{X}^{\mathcal{M}}].
% \end{equation}
% %
\textit{During inference}, the image generation starts with all masked tokens $\mathcal{X}^{\mathcal{M}} = [[\mathtt{MASK}], \cdots, [\mathtt{MASK}]]$, and gradually fills up the latent representation with scattered predictions in parallel.
Following MaskGIT~\cite{chang2022maskgit}, \system does not generate all tokens of the entire image at once,
since this process is inconsistent with our training procedure.
Instead, we infer the image tokens iteratively with the cosine masking schedule over $T$ iterations.
By default, we set $T$ to 50.

\vspace{-7pt}
\subsection{Pre-Training (PT)}
\label{sec:unigen_pretrain}
\vspace{-7pt}

The goal of pre-training is to develop \system's visual generation capability while preserving its potential for multimodal understanding.
Thus, we only optimize the generation projector and the LLM with other parameters frozen.
We also include image-to-text and text-only pre-training to keep \system's language modeling capability.
To encourage a better alignment between discrete image tokens and the text,
we directly use the generation encoder for understanding tasks \textit{but only} in this stage.
We empirically find that this design can significantly improve the image generation performance.
Specifically, we employ an ``easy-to-difficult'' strategy through a two-stage process.

\noindent\textbf{PT-1 Stage} seeks to align the image and text embeddings and predict the distribution of basic visual concepts.
We share the finding~\cite{wu2024janus} that the distribution of ImageNet~\cite{ridnik2021imagenet} can serve as an effective warm-up.
However, we propose that \textit{using image captions, rather than image category names, for text-to-image generation leads to better convergence}.
Therefore, we re-annotate the ImageNet dataset using Qwen2.5-VL-7B~\cite{bai2025qwen2}
and generate fine-grained captions for each image.
Similarly, we re-caption images from CC-3M~\cite{sharma2018conceptual}, CC-12M~\cite{changpinyo2021conceptual} and SAM-11M~\cite{kirillov2023segment}. These re-captioned datasets, along with ImageNet, form a 40M image-text pair corpus used for image-to-text pre-training. For text-only pre-training, we use RefinedWeb~\cite{penedo2023refinedweb}.

\noindent\textbf{PT-2 Stage} further facilitates \system to generalize to wider visual generation capabilities.
We augment ImageNet with our re-annotated CC-3M, CC-12M, and SAM-11M as PT-2 text-to-image dataset,
while using the same image-to-text and text-only ones.
We argue that \textit{training data with a richer distribution enables more accurate control over generation patterns}.
We name the model trained in this stage as \textbf{\systempt}.

\vspace{-2pt}
\subsection{Supervised Fine-Tuning (SFT)}
\label{sec:unigen_sft}
\vspace{-2pt}

In the SFT stage, \system is jointly trained on the image understanding and generation tasks.
We fine-tune the generation projectors, understanding projectors, and the LLM, while still keeping the vision encoders frozen.
\textbf{For image understanding}, we notice that the knowledge-centric understanding is limited during pre-training stages.
To enhance related capabilities, we adopt the strong image mixture from SlowFast-LLaVA-1.5~\cite{xu2025slowfast}, which was carefully curated from open-source datasets with 4.67M multimodal VQA samples.
\textbf{For image generation}, prior work~\cite{chen2025janus} uses high-quality synthetic data that can enable fast and robust training convergence. We share this observation by using the JourneyDB~\cite{sun2023journeydb} and text-2-image-2M~\cite{texttoimage2m2024} to improve the aesthetic quality of our generated images.
We name the model trained in this stage as \textbf{\systemsft}.

\subsection{Direct Preference Optimization (DPO)}
\label{sec:unigen_dpo}

\begin{figure*}[t]
    \centering
    \includegraphics[width=\linewidth]{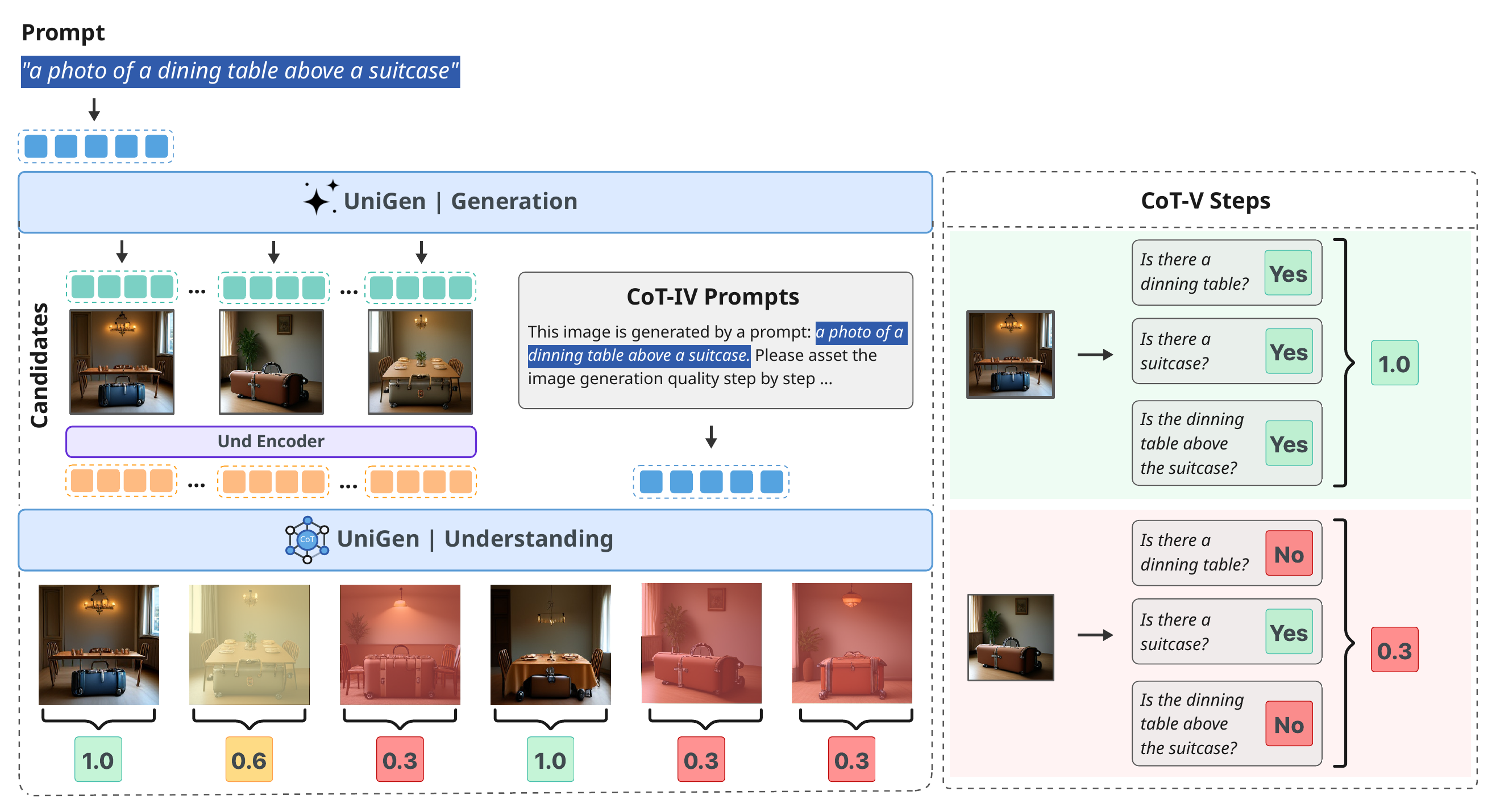}
    \vspace{-15pt}
    \caption{
        \textbf{The workflow of \system using test-time scaling and \cotshort.} \textbf{Left:} Illustration of \textit{Best-of-N} selection with \cotshort. \system first generates 6 image candidates and then selects the two images with the highest score after self-verification using \cotshort. \textbf{Right:} Visualization of the step-by-step reasoning process in \cotshort for computing the final quality score. }
    \vspace{-7pt}
    \label{fig:unigen_testtime}
\end{figure*}

We further enhance \system by aligning its outputs with human preference through DPO.
We first discuss how we construct our synthetic preference dataset, then describe our DPO algorithm.

\noindent \textbf{Preference Dataset.}
We leverage \systemsft to generate the images for our preference dataset. For a given prompt, 20 images are generated. A preferred and rejected sample pair is constructed by evaluating the coherence between each image and the prompt. To improve the data robustness, we cover short, medium and long prompts in our datasets.

\begin{itemize}[leftmargin=15pt]
    \item \textit{For short prompts}, we use the prompts from PARM~\cite{guo2025can}, which generally depict the objects, and their attributes and relationship in the scene. Following~\cite{guo2025can}, we use the \textsc{GenEval} metrics to evaluate the generation quality of the images. We select the highest-scored example as preferred and the lowest as rejected. This totally offers 6K pairs of image-text pairs with preference labels.
    \item \textit{For medium prompts}, we sample 6K from the T2I-Comp~\cite{huang2023t2i} training set, which feature more complex compositional concepts. We use Qwen2.5VL-7B to assess image-prompt consistency by decomposing each prompt into fine-grained visual questions. Each question receives a ``yes'' if the image aligns with the description, and ``no'' otherwise. The final consistency score $\mathcal{S}$ is averaged from these answers and used to select preferred and rejected image pairs.
    \item \textit{For long prompts}, we leverage 6K randomly sampled prompts from our re-annotated SA1B that contains high-quality images with rich semantic captions. Since it is challenging to evaluate the semantic alignment between an image and a long prompt, we follow the same procedure of medium prompts to label and select preference pairs.

\end{itemize}

\noindent \textbf{DPO Training.} We use the vanilla DPO training loss $\mathcal{L}_{DPO}$~\cite{rafailov2023direct}
\begin{equation}
    \small
    \mathcal{L}_{DPO} = -\mathbb{E}_{(x, y_w, y_l) \sim \mathcal{D}} \left[ \log \sigma \left( \beta \log \frac{\pi_{\text{DPO}}(y_w \mid x)}{\pi_{\text{SFT}}(y_w \mid x)} - \beta \log \frac{\pi_{\text{DPO}}(y_l \mid x)}{\pi_{\text{SFT}}(y_l \mid x)} \right) \right]
\end{equation}
where $\pi_{\text{SFT}}$ is our \systemsft model, $\pi_{\text{DPO}}$ is the optimizing \systemdpo model,
$y_w$ and $y_l$ are preferred and rejected examples for each prompt $x$, respectively,
and $\beta$ is a hyperparameter controlling the deviation from $\pi_{\text{DPO}}$ to $\pi_{\text{SFT}}$.
We only optimize the generation modules (\ie, freezing the understanding encoder and projector) in this stage.
The training ends in one epoch with a batch size of 64 and a learning rate of $1e^{-5}$. 
We empirically find that \textit{this DPO training does not impair \system's understanding performance.}
We name the model trained in this stage as \textbf{\systemdpo}.

\begin{figure*}[t]
    \centering
    \includegraphics[width=\linewidth]{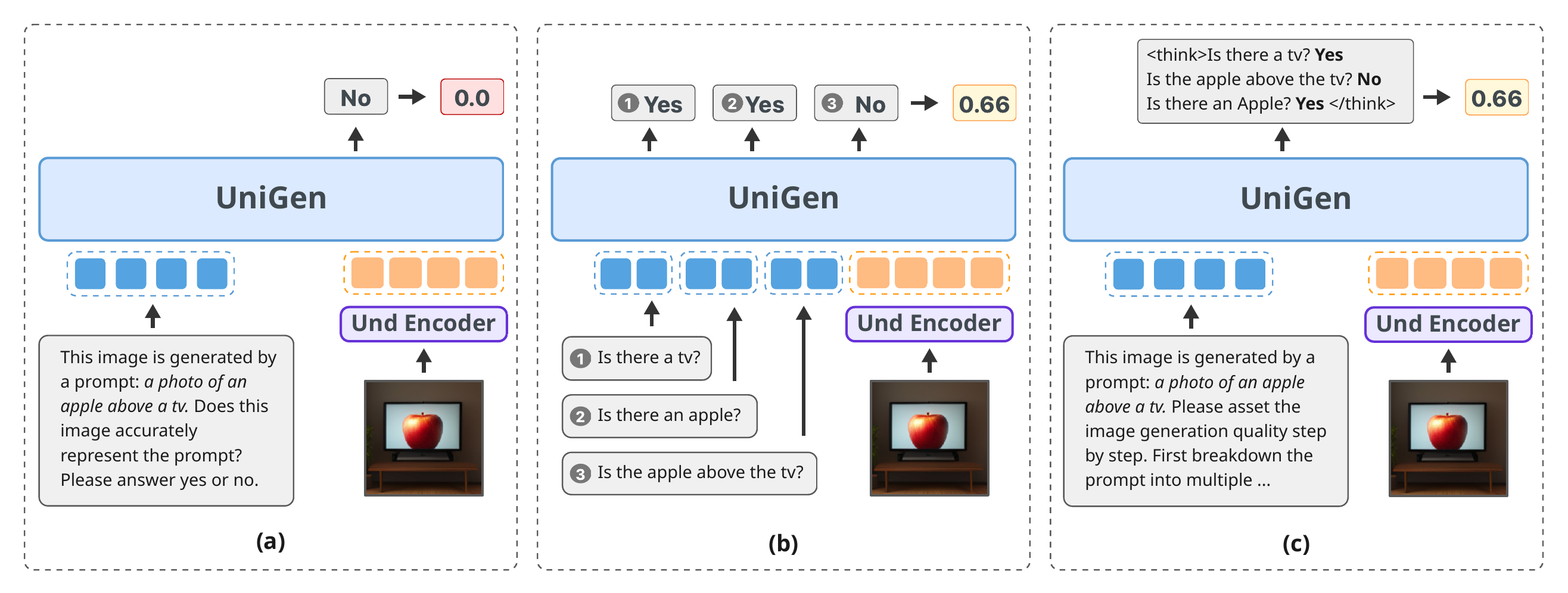}
    \vspace{-15pt}
    \caption{
        \textbf{An example of using different image verification methods}: \textbf{(a)} Outcome Verification, \textbf{(b)} Rule-based Verification and \textbf{(c)} Chain-of-Thought Verification.
    }
    \vspace{-5pt}
    \label{fig:verification_methods}
\end{figure*}

\vspace{-5pt}
\subsection{Test-Time Scaling}
\label{sec:unigen_testtime}
\vspace{-5pt}

Recent studies have shown the effectiveness of test-time scaling on improving both image understanding~\cite{wang2025visualprm,zhu2025internvl3} and generation~\cite{guo2025can}.
We employ the Best-of-N evaluation strategy
and leverage \system's understanding ability to conduct self-critique for image generation verification.
The general workflow is illustrated in Fig.~\ref{fig:unigen_testtime}.
\textit{First}, \system generates $N$ candidate images for a given prompt.
\textit{Second}, we input each generated image along with its prompt into \system, which evaluates the alignment between the image and its textual description by returning a quality score $\mathcal{S}$.
\textit{Third}, we select the top-$K$ images with the highest scores as our final results.

In this work, we propose three verification methods as shown in Fig.~\ref{fig:verification_methods}.

\begin{itemize}[leftmargin=15pt]
    \item \textbf{Outcome Verification (OV)} simply prompts \system to directly judge the coherence of the input prompt and each image candidate, giving a binary score (\ie ``yes'' for a good match and ``no'' for a failure generation). We randomly select one if there are candidates with the same score.
    
    \item \textbf{Rule-based Verification (RV)} breaks down each prompt into several atomic questions based on pre-defined rules, then sequentially feeds them with the generated image into \system for quality verification. The results of all sub-questions are averaged as the final quality score.

    \item \textbf{Chain-of-Thought Verification (CoT-V)} instructs the model to think step-by-step and verifies each atomic fact according to the prompt and each generated image, following the CoT format:
    \texttt{<think\_start>}$\mathtt{Q_1}$\texttt{?} $\mathtt{A_1}$\texttt{;} $\mathtt{\cdots}$ $\mathtt{Q_n}$? $\mathtt{A_n}$\texttt{;}\texttt{<think\_end>}.
    % \texttt{<answer\_start>}$\mathtt{\bar{A}}$\texttt{<answer\_end>}.
    We compute the final quality score $\mathcal{S}$ by parsing the CoT outputs.
    Specifically, given a text prompt $T$ and a generated image $I$,
    \cotshort produces a list of visual questions $Q = \{Q_1, \cdots, Q_n\}$
    and their corresponding the answers $A = \{A_1, \cdots, A_n\}$.
    The final score is defined as:
        \begin{equation}
            \small
            \mathcal{S}(T, I) = \frac{1}{n} \sum_{j=1}^n s_j(T, I), \; \text{where} \;
            s_j(T, I) = 
            \begin{cases}
            1, & \text{if } A_j = \text{``yes''} \\
            0, & \text{otherwise}
            \end{cases}.
        \end{equation}
\end{itemize}

OV relies on \system's pattern-matching capabilities without intermediate reasoning.
RV incorporates a rule-driven reasoning process into test-time scaling.
Although effective on well-structured prompts,
RV struggles with free-form or complex instructions, such as those in DPG-Bench~\cite{hu2024ella}.
\cotshort leverages the strengths of both approaches,
enabling reasoning-driven image verification without the need for manual prompt decomposition.
Thus, we use \cotshort as our default verification method.

\vspace{-5pt}
\subsubsection{\cotshort Post-Training}
\label{sec:unigen_cotv}
\vspace{-3pt}

\system has not been precisely trained to generate CoT responses.
Here we introduce a lightweight post-training strategy upon \systemdpo,
equipping it with the ability of CoT-based verification.

\noindent \textbf{Data.} To construct the \cotshort post-training data,
we reuse the image-text pairs collected during the DPO stage (Sec.~\ref{sec:unigen_dpo}).
For prompts sourced from PARM,
we extract the question-answer pairs via rule-based matching, since they are built upon a clear structure~\cite{ghosh2023geneval}. 
For prompts from T2I-Comp that are more complicated,
we first guide Qwen2.5-7B~\cite{yang2024qwen2} to generate a series of atomic questions, then query Qwen2.5-7B-VL with each image-question pair to obtain their binary pseudo labels. 
We exclude the prompts from SA-1B due to the lower quality of decomposed visual questions. We empirically find that most of the decomposed questions do not fully cover the visual concepts of the original caption.
We totally sample 20K image-question-answer triplets from both prompt sources. 

\noindent \textbf{Training.}
We format the above 20K training pairs as instruction-following conversations,
and feed them into \systemdpo for supervised fine-tuning.
In this stage, we only optimize the understanding projector and the LLM.
To ensure not impairing \system's general understanding capabilities,
we fine-tune \system on this \cotshort dataset for only $500$ steps using a small learning rate of $1 \times 10^{-5}$.
The model trained after this stage is our final model and we name it as \textbf{\system}.

%% file: sec/4_experiments.tex
\vspace{-5pt}
\section{Experiments}
\label{sec:experiments}
\vspace{-5pt}

\begin{figure*}[t]
    \centering
    \includegraphics[width=\linewidth]{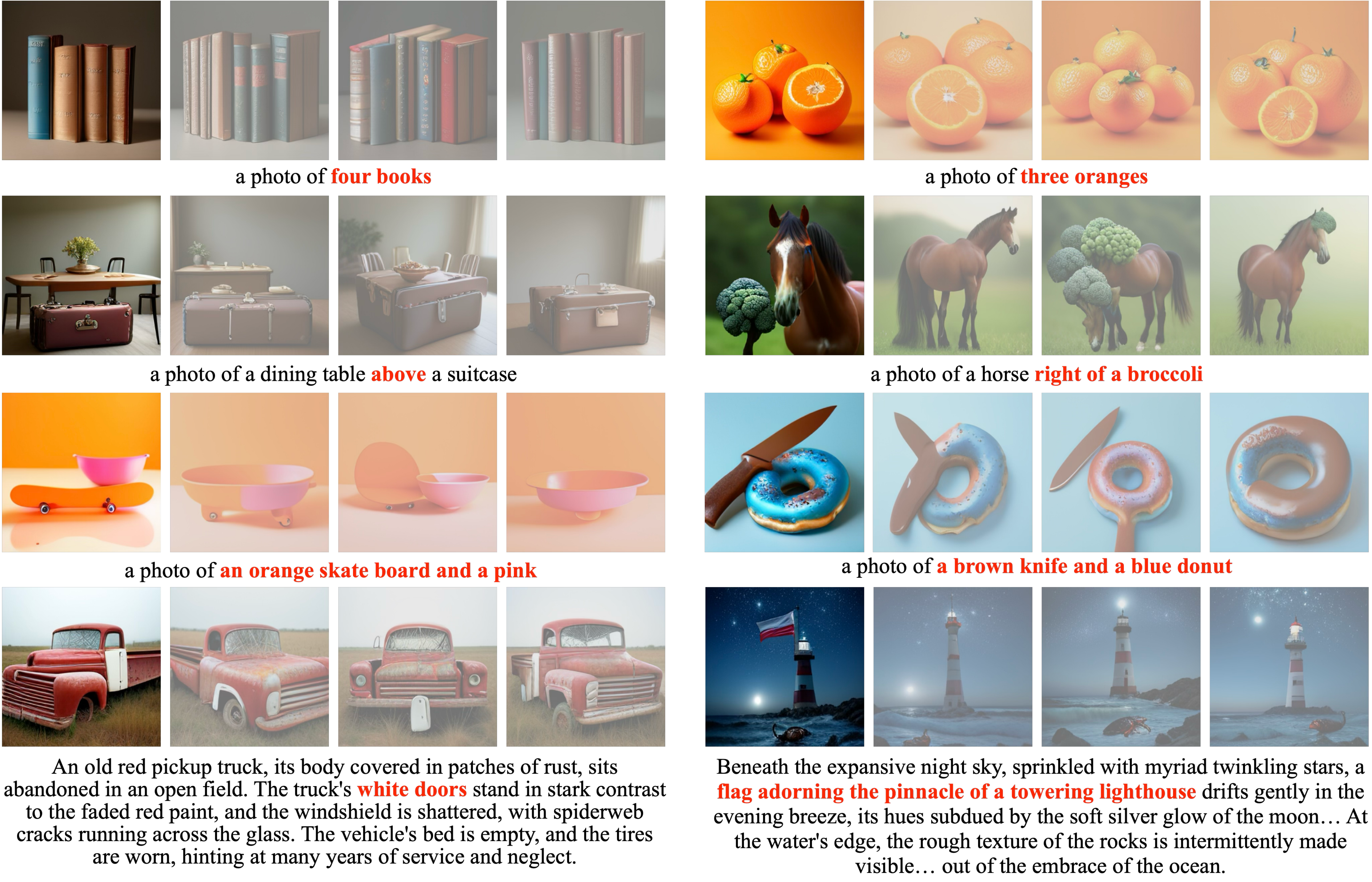}
    \vspace{-0.2in}
    \caption{\textbf{Visual examples of \system's results using \cotshort.} The first three rows show examples for counting, position, and color attribute, respectively, and the last row shows images generated by free-form prompts. The first column contains images selected by \system as the test-time verifier.}
    \label{fig:cot_res}
    \vspace{-0.1in}
\end{figure*}

\subsection{Implementation Details}
\label{sec:implementation_details}
\vspace{-3pt}

\input{tables/understanding_final}

We use 32 H100-80G GPUs for pre-training stages and 8 H100-80G GPUs for the others.
\system is built upon the pre-trained Qwen2.5-1.5B~\cite{yang2024qwen2}.
We adopt MAGVITv2 from Show-o~\cite{xie2024show} as our discrete visual encoder with input resolution of $256\times256$
and SigLIP~\cite{zhai2023sigmoid} as our continuous visual encoder.
As discussed in Sec~\ref{sec:unigen_arch}, we use MAGVITv2 for both understanding and generation in PT-1 and PT-2, and keep using SigLIP as the understanding encoder after SFT.

\textbf{Training.} To perform masked token prediction for image generation task, we follow Show-o~\cite{xie2024show} to use bidirectional attention mask within image tokens,
but keeps the causality within text tokens and between multimodal tokens.
The same attention masking strategy is adopted for the image understanding task.
For text-only data, we use the causal attention mechanism.
Detailed hyper-parameters, such as batch size and learning rate, for each training stage are described in Appendix Table~\ref{tab:hyperp_train} with more training details in Appendix Sec.~\ref{sec:train_data_supp}.

\textbf{Inference and Evaluation.} We follow the common practice of image generation to use classifier-free guidance~\cite{ho2022classifier} and set the scale to $5.0$.
In addition, we follow MaskGIT~\cite{chang2022maskgit} to adopt the cosine masking scheduler in inference and set the default number of steps to $T=50$. We use MAGVITv2 decoder to project the visual tokens back to the pixel space. For test-time scaling with \cotshort, we generate $N=20$ image candidates per text prompt and select top-K ($K=4$) out of it, sending for evaluation on \textsc{GenEval} and \textsc{DPG-bench}.

\vspace{-5pt}
\subsection{Main Results}
\label{sec:main_results}
\vspace{-3pt}

We report the performance of \system on various image understanding and generation benchmarks (details are discussed in Appendix Sec.~\ref{sec:benchmarks_supp}) and show some qualitative images in Fig~\ref{fig:cot_res}.
We mainly compare \system with state-of-the-art unified LLMs in Table~\ref{tab:mmu_cmp} and \ref{tab:t2i_sota_cmp}, but also reference strong specialist models to 
understand our position in the whole picture of MLLMs. Here we highlight the following observation.

\noindent\textbf{First, \system achieves state-of-the-art results across understanding benchmarks compared to existing unified MLLMs.}
Specifically, \system outperforms Janus-Pro on RealWorld-QA, AI2D and MathVista by $+5.6\%$, $+3.7\%$, and $+7.8\%$, respectively. We believe our improvements are mainly driven by using \textit{(i)} the decoupled generation and understanding encoders and \textit{(ii)} the stronger SFT data mixture. Notably, \system is even comparable with some strong understanding-only MLLMs, such as LLaVA-OV-0.5B and MM1.5-1B, even though they use much higher input resolutions.

\noindent\textbf{Second, \system significantly outperforms existing unified MLLMs and strong generation-only models on text-to-image benchmarks.}
Using \textsc{GenEval} in Table~\ref{tab:t2i_sota_cmp} as an example, \system achieves the overall score of $0.78$, significantly outperforming Janus-Pro by $0.05$.
Besides, our model demonstrates an overwhelming advantage on the ``Counting'' task by $+0.27$ higher than Janus-Pro.
\system even beats a range of superior generation-only models (\eg, outperforming DALLE-2, and Emu3 by $+0.26$, and $+0.24$, respectively), even though they are with much larger model sizes.
Similarly, \system outperforms existing models by a clear margin on \textsc{DPG-bench} as shown in Table~\ref{tab:t2i_sota_cmp},
outperforming Show-o and Janus-Pro by $+13.49$ and $+2.56$, respectively.

\input{tables/t2i_bench}

\subsection{Ablation Studies}
\label{sec:sec_ablation} 
\vspace{-5pt}

We first present the detailed impact of each stage, then present our studies of DPO and Test-Time Scaling stages.
Please refer to Appendix Sec.~\ref{sec:more_ablation_supp} for more ablations on PT-1, PT-2, and SFT.

\vspace{-5pt}
\subsubsection{Impact of Different Training Stages}
\label{sec:ablation_overall}
\vspace{-2pt}

\input{tables/stage_ablation_final}

We examine our training pipeline by showing the understanding and generation performance after each stage in Table~\ref{tab:stage_ablation_final}. Here we highlight some key observations.

\noindent\textbf{First, \system's generation performance is consistently improved across stages} as indicated by the increasing numbers of \textsc{GenEval} and \textsc{DPG-bench}.
The pre-training stages aim to warm up the generation capability of \system. The SFT boosts the \textsc{GenEval} and \textsc{DPG-bench} by using high-quality generation datasets. With the effectiveness of our preference data, the DPO stage significantly improves \textsc{GenEval} and \textsc{DPG-bench} to $0.73$ ($+0.10$) and $84.89$ ($+2.14$), respectively. \cotshort further enhances the scores to $0.78$ ($+0.05$) and $85.19$ ($+0.3$) via test-time scaling.

\noindent\textbf{Second, \system's strong understanding capability is stimulated in the SFT stage and can be maintained in the following stages.} The SFT stage promotes the instruction following capability of \system that leads to strong performance on understanding benchmarks. We use DPO for aligning \system's output with preference data for image generation and observe that this stage successfully maintains the strong understanding capability. \cotshort contains an additional lightweight fine-tuning to encourage the CoT verification during test-time scaling. The results show that it does not sacrifice the general understanding capability, except for a slight regression on RealWorld-QA.
We attribute this regression to the distribution gap between \cotshort's synthetic training data and the real world images in RealWorld-QA.

\vspace{-5pt}
\subsubsection{Ablation of \cotshort}
\vspace{-2pt}

Here we evaluate different verification methods discussed in Sec.~\ref{sec:unigen_testtime} with the following highlights.

\input{tables/testtime_scale_ablation}

\noindent\textbf{First, CoT verification achieves the best performance and prompting \system's thinking process is important.}
As shown in \Cref{tab:testtime_scale_ablation}, using \textit{Outcome verification} shows no improvement, while using CoT thinking obtains significant boost of generation performance on both \textsc{GenEval} and \textsc{DPG-bench}. We also observe that \emph{Rule-based verification} is also effective that leads to $0.75$ on \textsc{GenEval}. However, it is not general enough to be used on free-form prompts. Comparing the results from \textit{CoT Verification} and \textit{Rule-based Verification}, we can see that prompting the model itself to think is beneficial for more reliable critique.

\noindent\textbf{Second, \cotshort post-training is essential for strong test-time verification.} As shown in Table~\ref{tab:cot_ablation},
directly using \system without \cotshort post-train leads to notable performance drop, especially for \textsc{GenEval}. This comparison demonstrates that \cotshort post-train is pivotal for CoT verification.

\noindent\textbf{Third, \cotshort can effectively generalize to other models.}
We fine-tune Show-o with DPO and \cotshort with our generated data to boost its generation performance. Results in \Cref{tab:cot_ablation} show that \cotshort is a general technique that can also enhance Show-o's generation performance.

\vspace{-2pt}
\subsubsection{Ablation of DPO}
\input{tables/dpo_ablation}

\vspace{-2pt}

We ablate the contribution of each data source and demonstrate the effectiveness of our DPO data on other unified models.

\noindent\textbf{First, every prompt source contributes positively to generation performance.} Table~\ref{tab:dpo_ablation} shows that adding only PARM DPO data results in remarkable improvements (row1 vs. row2), while further adding T2I-Comp mainly benefits \textsc{DPG-bench} (row2 vs. row3). \systemdpo with all of the prompts, introduces the best overall performance (row3 vs. row4). 

\noindent\textbf{Second, our DPO data also largely improves Show-o, showing that it is generalizable to other unified models.} When fine-tuning Show-o directly with our DPO data, we also observe notable gain, from $0.56$ to $0.64$ on \textsc{GenEval} and from $71.70$ to $76.32$ on \textsc{DPG-bench} as shown in Table~\ref{tab:dpo_ablation}.

%% file: tables/understanding_final.tex
\begingroup
    \begin{table*}[t]
    \caption{
        \textbf{Comparison with state-of-the-art models on image understanding benchmarks.}
        $^*$denotes reproduced results and RW-QA denotes RealWorld-QA.
    }
    \centering
    \footnotesize
    
    \ra{1}
    \addtolength{\tabcolsep}{-2pt} 
    \resizebox{\linewidth}{!}{
    \begin{tabular}{lcc|cccccccc}
    \toprule
    \textbf{Model} & \textbf{\#Params} & \textbf{Res.} &\textbf{AI2D} &\textbf{GQA} &\textbf{POPE} &\textbf{MMMU} &\textbf{MathVista} & \textbf{RW-QA} & \textbf{ScienceQA}& \textbf{Seedbench} \\
    \cmidrule{1-2}\cmidrule{3-11}
    \multicolumn{11}{c}{\textit{Understanding MLLMs}}\\
    \midrule
    LLaVA-OV~\cite{li2024llava} & 0.5B& AnyRes & 57.1 &- & - & 31.4 & 34.8 & 55.6 & 67.2 & 65.5\\
    MM1.5~\cite{zhang2024mm1} & 1B & AnyRes & 59.3 & - &88.1 &  35.8 &37.2 & 53.3& 82.1 & 70.2\\
    LLaVA 1.5~\cite{liu2024improved} & 7B&336  & \;\,55.1* & 	62.0& 86.1	&\;\,36.3*& \;\,26.7*& \;\,55.8* &66.8	&66.1 \\
     \midrule
    \multicolumn{11}{c}{\textit{Unified MLLMs}}\\
    \midrule
    Show-o~\cite{xie2024show} &  1.3B& 336 & \;\,36.2* & \;\,61.0* & 84.5 & 27.4 & \;\,22.1* & \;\,48.5* &  \;\,42.7* & \;\,61.5* \\
    Janus~\cite{wu2024janus} &1.3B& 384 &\;\,49.0* & 59.1 & 87.0 & 30.5 & \;\,33.7* & \;\,48.4* & \;\,76.5* & 63.7 \\
    Janus-Pro~\cite {chen2025janus} &1.5B& 384 & \;\,63.7* & 59.3 & 86.2 & \textbf{36.3} & \;\,36.8* & \;\,51.1* & \;\,75.5* & 68.3 \\
    Vila-U~\cite{wu2024vila} &7B & 384 & - & 60.8 & 85.8  & - & - & - & -& 56.3\\
    MMAR~\cite{yang2024mmar} &7B & 256 &- &-& 83.0& -&-& -& - &64.5 \\
    UniToken~\cite{jiao2025unitoken} & 7B  & 384& \textbf{68.7} & - & - & 32.8 & 38.5 & - & -&69.9 \\
    \rowcolor{lightblue} \system & 1.5B & 384 & 67.4 & \textbf{62.3} & \textbf{87.8} & 32.3 & \textbf{44.6} & \textbf{56.7} & \textbf{79.4} & \textbf{70.8}\\
    \midrule
    \end{tabular}
    }
    \label{tab:mmu_cmp}
\vspace{-0.1in}
\end{table*}
\endgroup

%% file: tables/t2i_bench.tex
\begingroup
    \begin{table*}[t]
    
    \small
    \centering
    \caption{
  \textbf{Comparison with state-of-the-art models on \textsc{GenEval} and \textsc{DPG-Bench} benchmark.}
    }
    \vspace{-5pt}
    \footnotesize
    \addtolength{\tabcolsep}{-2pt}
    \ra{1}
    \resizebox{\linewidth}{!}{%
    \begin{tabular}{lccccccccc}
    \toprule
    \multirow{2}{*}{\textbf{Model}} & \multirow{2}{*}{\textbf{\#~Params}} & \multicolumn{5}{c}{\textbf{GenEval$\uparrow$}} & \multicolumn{3}{c}{\textbf{DPG-Bench$\uparrow$}} \\
    \cmidrule(l){3-7}\cmidrule(l){8-10} 
    & & \textbf{Two Obj.}  & \textbf{Counting} & \textbf{Position} & \textbf{Color Attri.} & \textbf{Overall}  & \textbf{Global}& \textbf{Relation}&  \textbf{Overall}\\
    % \cmidrule{1-2}\cmidrule{3-9}
    \midrule
    \multicolumn{10}{c}{\textit{Text-to-Image Generation Models}}\\
    \midrule
    DALLE-2~\cite{ramesh2022hierarchical} & 6.5B & 0.66 & 0.49 & 0.10 & 0.19 & 0.52 & - &  - & -\\
    
    DALLE-3\cite{betker2023improving} & - & 0.87 & 0.47 & 0.43 & 0.45 & 0.67 & 90.97 & 90.58 & 83.50 \\
    Emu3~\cite{wang2024emu3} & 8B & 0.71 & 0.34 & 0.17 & 0.21 & 0.54 & 85.21 & 90.22 & 80.60\\ 
    SDXL~\cite{podellsdxl} & 2.6B  &0.74 &0.39 &0.15 &0.23 &0.55 & 83.27 & 86.76 &74.65  \\
    SimpleAR~\cite{wang2025simplear} & 1.5B &0.90  &- &0.28 &0.45 &0.63 & 87.97 & 88.33 & 81.97\\
    Infinity~\cite{han2024infinity} & 2B & 0.85 & - & 0.49 & 0.57 & 0.73 &93.11 & 90.76 & 83.46 \\
    \midrule 
    \multicolumn{10}{c}{\textit{Unified MLLMs}}\\
    \midrule
Show-o~\cite{xie2024show}& 1.3B  & 0.52 & 0.49 & 0.11 & 0.28 & 0.53 & \;\,80.39* & \;\,83.36* & \;\,71.70*\\
 Janus~\cite{wu2024janus} & 1.3B& 0.68 & 0.30 & 0.46 & 0.42 & 0.61& 82.33 & 85.46 & 79.68 \\
 Janus-Pro~\cite{chen2025janus} & 1.5B& 0.82 & 0.51 & 0.65 & 0.56 & 0.73 &87.58 & 88.98 & 82.63 \\
 ILLUME~\cite{wang2024illume} & 7B  & 0.86 & 0.45 & 0.39 & 0.28 & 0.61 &- &- &- \\
 UniToken~\cite{jiao2025unitoken} & 7B & 0.80  &0.35  &0.38  &0.39 & 0.63 &- &- &- \\
 VARGPT-v1.1~\cite{zhuang2025vargpt} & 9B &0.53 &0.48 &0.13 &0.21 &0.53 &84.83 & 88.13 & 78.59\\
 TokenFlow-XL~\cite{qu2024tokenflow} & 13B & 0.72 & 0.45 & 0.45 & 0.42 & 0.63 &78.72 &85.22 & 73.38 \\
 \rowcolor{lightblue} \system  & 1.5B & 0.94& 0.78 & 0.57 &0.54 &\textbf{0.78} &91.95 & 92.04 & \textbf{85.19} \\
    \midrule
    \end{tabular}
    }
    \vspace{-5pt}
    \label{tab:t2i_sota_cmp}
\end{table*}
\endgroup

%% file: tables/stage_ablation_final.tex
\begingroup
    \begin{table*}[t]
    \caption{
        \textbf{Ablation of different stages of our model on image understanding benchmarks.}
    }
    \vspace{-5pt}
    \centering
    \scriptsize
    \addtolength{\tabcolsep}{1pt}
    \ra{1.2}
    \addtolength{\tabcolsep}{-5pt} 
    \resizebox{\linewidth}{!}{
    \begin{tabular}{l|c|cc|cccccccc}
    \toprule
    \textbf{Model} & \textbf{Stage}& \textbf{GenEval} & \textbf{DPG-Bench} &\textbf{AI2D} &\textbf{GQA} &\textbf{POPE} &\textbf{MMMU} &\textbf{MathVista} & \textbf{RW-QA} & \textbf{ScienceQA}& \textbf{Seedbench} \\
    \midrule
    \multirow{5}{*}{\system} & PT-1 &0.53 & 78.14& - & -& -& -& -& -& -& - \\
    & PT-2 &0.55 & 80.71& - & -& -& -& -& -& -& - \\
    & SFT & 0.63 &82.75 &68.0 &62.5 &87.4 &32.4 &45.2 &58.6 &79.7 & 71.1\\
     & DPO & 0.73 & 84.89& 67.9	 &62.4 &88.0& 32.9& 45.0& 59.0&79.5&71.0\\
     & \cotshort & 0.78 & 85.19& \multirow{1}{*}{67.4} & \multirow{1}{*}{62.3} & \multirow{1}{*}{87.8}  & \multirow{1}{*}{32.3} & \multirow{1}{*}{44.6} & \multirow{1}{*}{56.7} & \multirow{1}{*}{79.4} & \multirow{1}{*}{70.8} \\
    \bottomrule
    \end{tabular}}
    \label{tab:stage_ablation_final}
\vspace{-0.15in}
\end{table*}
\endgroup

%% file: tables/testtime_scale_ablation.tex
\begin{table}[t]
\centering
% \caption{Ablation studies. Left: \cotshort; Right: DPO.}

\begin{minipage}[t]{0.52\linewidth}
\scriptsize
\centering
\addtolength{\tabcolsep}{-2.5pt}
\ra{1.1}
\caption{\textbf{Ablation of verification methods.}}
\resizebox{\linewidth}{!}{%
\begin{tabular}{l|ccccc}
\toprule
Method & Outcome & Rule& CoT& GenEval & DPG-Bench \\
\midrule
\multirow{4}{*}{\system} &\ding{55} & \ding{55} & \ding{55} & 0.74& 85.02\\
&\ding{51} & \ding{55} & \ding{55} & 0.74& 85.00\\
 &\ding{55} & \ding{51} & \ding{55} &0.75 & - \\
  &\ding{55} & \ding{55} & \ding{51} & 0.78 & 85.19\\

\bottomrule
\end{tabular}
\label{tab:testtime_scale_ablation}
}
\end{minipage}
\hfill
\begin{minipage}[t]{0.45\linewidth}
\scriptsize
\centering
\caption{\textbf{Ablation of \cotshort post-training.}}
\addtolength{\tabcolsep}{-3pt}
\ra{1.2}
\resizebox{\linewidth}{!}{%
\begin{tabular}{lc|cc}
\toprule
Method &\cotshort Post-train & \textsc{GenEval} & \textsc{DPG-bench} \\
\midrule
\multirow{2}{*}{Show-o}&\ding{55} &  0.64 &76.32 \\
&\ding{51} &  0.66 & 77.09\\
\midrule
\multirow{2}{*}{\system} &\ding{55} &0.74 & 84.89\\
&\ding{51} & 0.78&  85.19\\
\bottomrule
\end{tabular}
\label{tab:cot_ablation}
}
\end{minipage}
\vspace{-10pt}

\end{table}

%% file: tables/dpo_ablation.tex
\begin{wraptable}{r}{0.5\textwidth} 
\scriptsize
\centering
\vspace{-0.2in}
\caption{\textbf{Ablation study of DPO.} The results are from \systemdpo without test-time scaling.}
\vspace{-0.1in}
\addtolength{\tabcolsep}{-3pt}
 \ra{1.1}
\resizebox{\linewidth}{!}
{\begin{tabular}{l|ccc|cc}
\toprule
Method & PARM & T2I-Comp& SA1B& GenEval & DPG-bench \\
\midrule
\multirow{4}{*}{\system} &\ding{55} & \ding{55} & \ding{55} & 0.63 & 82.75 \\
&\ding{51} & \ding{55} & \ding{55} & 0.73 & 83.48\\
 &\ding{51} & \ding{51} & \ding{55} & 0.72 & 84.09 \\
  &\ding{51} & \ding{51} & \ding{51} & 0.74 & 84.89 \\
\midrule
\multirow{2}{*}{Show-o} &\ding{55} & \ding{55} & \ding{55} & 0.56 & 71.70
\\
& \ding{51} & \ding{51} & \ding{51} & 0.64& 76.32
\\
\bottomrule
\end{tabular}
}
\vspace{-0.1in}
\label{tab:dpo_ablation}
\end{wraptable}

%% file: sec/6_conclusion.tex
\vspace{-2pt}
\section{Conclusion}
\label{sec:conclusion}
\vspace{-3pt}

We present \system, an MLLM for unified multimodal understanding and generation. We discuss the key factors along the entire training pipeline and propose optimization methods to improve the performance.
We also make the first attempt to collaborate \system's understanding and generation capabilities,
by enabling \system to perform as both image generator and verifier during test-time scaling.
As a result, we successfully further boost the image generation quality by a clear margin.
Trained with only open-source datasets, \system achieves the state-of-the-art performance across extensive understanding and generation benchmarks.
We hope our exploration and ablation studies provide insights to the future development of strong unified MLLMs.

\textbf{Limitation.} \textit{First}, we instantiate \system with only a 1.5B model, since larger scales will impose much higher demands on the computational cost. However, larger models have been shown effective for improving both understanding and generation performance~\cite{chen2025janus}.
\textit{Second}, our generation capability targets at promoting semantic alignment between the input text prompt and the generated image, therefore we only focus on a resolution of $256\times256$. We plan to support higher resolution image generation, such as 480p or even 1080p, which is valuable for improving the visual fidelity.
\textit{Third}, although achieving convincing results on DPG-Bench, \cotshort is still limited for complicated text prompt, due to the noisy CoT data generated by Qwen2.5VL as a pseudo labeler. This could be largely relieved by using a stronger pseudo labeler or leveraging human filtering in the future. Equipping \system with stronger reasoning and CoT capabilities in an earlier stage is also a promising direction.

\subsection*{Acknowledgment}

We thank Haiming Gang, Jesse Allardice, Shiyu Li, and Yifan Jiang for their kind help.

%% file: sec/X_supple.tex
\section{Benchmarks and Evaluation Protocol}
\label{sec:benchmarks_supp}

\textbf{For image understanding}, we include widely-used \textit{(i)} general VQA benchmarks, such as GQA~\cite{hudson2019gqa}, RealWorld-QA~\cite{rwqa}, and Seedbench \cite{li2023seed},
\textit{(ii)} knowledge-based benchmarks, such as AI2D~\cite{kembhavi2016diagram}, MMMU~\cite{yue2023mmmu}, and MathVista~\cite{lumathvista},
and \textit{(iii)} hallucination benchmarks, such as POPE~\cite{Li-hallucination-2023}.
We leverage the \texttt{lmms-eval}\footnote{https://github.com/EvolvingLMMs-Lab/lmms-eval} toolkit to compute the results for the above benchmarks.

\textbf{For text-to-image generation benchmarks}, we report results on \textsc{GenEval}~\cite{ghosh2023geneval} and \textsc{DPG-bench}~\cite{hu2024ella} to comprehensively evaluate the semantic alignment between a text prompt and the generated images. To fairly compare with recent unified MLLMs~\cite{wu2024janus,chen2025janus,xie2024show}, our results are obtained using the official evaluation repository of \textsc{GenEval}\footnote{https://github.com/djghosh13/geneval/tree/main} and \textsc{DPG-bench}\footnote{https://github.com/TencentQQGYLab/ELLA/tree/main}.

\section{More Ablation Studies}
\label{sec:more_ablation_supp}

\subsection{Ablation of SFT}
\input{tables/sft_ablation}

By default, we use the image mixture from SlowFast-LLaVA-1.5~\cite{xu2025slowfast} as understanding datasets and JourneyDB and text-2-image-2M as the generation datasets. In this section, we ablate the datasets in Table~\ref{tab:sft_ablation} to evaluate their impacts and draw the following conclusion.

\noindent\textbf{First, using high-quality generation data is necessary for further lifting generation results.} JourneyDB and text-2-image-2M have much higher quality compared to the generation data used during the PT-2 stage. Table~\ref{tab:sft_ablation} (row1 vs. row2) shows that using high-quality generation data in the SFT stage results in better image generation performance.

\noindent\textbf{Second, using a stronger data mixture is crucial to improve the understanding performance, which is also helpful for fine-grained text-to-image generation.} As shown in Table~\ref{tab:sft_ablation} (row2 vs. row3), replacing SlowFast-LLaVA-1.5 mixture with LLaVA1.5's induces much worse understanding performance. Also, training with SlowFast-LLaVA-1.5 data produces higher results on \textsc{DPG-bench}. We believe a better understanding capability is important for comprehending the complex text prompts of \textsc{DPG-bench} that can eventually be beneficial for better text-to-image generation.

\subsection{Ablation of PT-1 and PT-2}

We explore the necessity and key factors of \system's pre-training stages.
We first discuss whether we should include the understanding dataset in the pre-training stages as shown in \Cref{tab:pretrain_und_data_ablation}. Second, we ablate the impact of the generation datasets to both generation and understanding performance in \Cref{tab:pretrain1_ablation} (for PT-1) and \Cref{tab:pretrain2_ablation} (for PT-2).
Since PT-1 and PT-2 are early stages in \system's training pipeline,
we continue the training to the SFT stage to verify their impact on the final performance more reliably.
Unless noted otherwise, all ablations in this section use \system's default SFT settings.
Here we highlight the following observations.

\noindent\textbf{First, including understanding data in pre-training stages is crucial for both generation and understanding performance.} In \Cref{tab:pretrain_und_data_ablation}'s row 1, we keep the default setting and only remove the understanding loss from the training objectives. We observe a significant performance decrease across generation and understanding benchmarks at the SFT stage. We attribute this to the fact that understanding data is important for a better vision-language alignment in early training stages, which is helpful for both image-to-text and text-to-image tasks.

\input{tables/pretrain_und_data_ablation}

\input{tables/pretrain1_ablation}
\input{tables/pretrain2_ablation}

\noindent\textbf{Second, the high-quality text-to-image task is more effective than the de facto class-to-image task in PT-1.} One common practice of unified MLLMs for pre-training is using the class-to-image task with ImageNet~\cite{wu2024janus,xie2024show}. However, we find that using ImageNet with fine-grained captions leads to better performance for understanding tasks in the \systemsft stage as shown in \Cref{tab:pretrain1_ablation} (row2 vs. row3). This is a result of better vision-language alignment introduced by the detailed caption-to-image mapping.

\noindent\textbf{Third, to maintain high performance on generation, we need both PT-1 and PT-2.} According to \Cref{tab:pretrain1_ablation} (row1 vs. row3) and \Cref{tab:pretrain2_ablation} (row1 vs. row6), we notice that completely removing PT-1 or PT-2 stage will largely decrease the generation metrics. Especially, eliminating PT-2 has a much bigger negative impact, leading to a dramatic drop of numbers on both \textsc{GenEval} and \textsc{DPG-bench}. Excluding PT-1 will not apparently affect \textsc{GenEval}, but hurts \textsc{DPG-bench}. This is because the prompts of \textsc{DPG-bench} are more complicated, thus more pre-training helps our model to better comprehend their semantics.

\noindent\textbf{Fourth, to keep a strong understanding performance, we need at least one of the PT-1 and PT-2.} According to \Cref{tab:pretrain_und_data_ablation}, we infer that the understanding performance will be destroyed if we remove both PT-1 and PT-2. However, discarding PT-1 or PT-2 in \Cref{tab:pretrain1_ablation} and \Cref{tab:pretrain2_ablation} does not impact understanding numbers. As a result, we recommend keeping at least one of them for good understanding capability and leveraging both of them for the best generation and understanding performance if the compute budget allows.

\noindent\textbf{Fifth, using high-quality captions in PT-2 is important for understanding and generation performance.} \Cref{tab:pretrain2_ablation} (row2 vs.~row6) demonstrates that using high-quality image captions results in stronger performance in both understanding and generation tasks. This is due to the better text-to-image and image-to-text alignment learned from the fine-grained captions.

\noindent\textbf{Sixth, each data source of PT-2 has meaningful contributions.} We remove each data component from the training set of PT-2 and observe that retaining all of them leads to the best performance as shown in \Cref{tab:pretrain2_ablation} (row3 to row6). This finding supports the usefulness of each dataset we curated.

\section{More Results}
\label{sec:more_results_supp}

We present the breakdown comparison of \system against state-of-the-art models on \textsc{GenEval} and \textsc{DPG-bench} in \Cref{tab:geneval_cmp} and \Cref{tab:dpg_cmp}.

\input{tables/geneval_final}
\input{tables/dpg_final}

\section{Details of Test-Time Strategies}
\label{sec:testtime_supp}

\vspace{-5pt}
\subsubsection{Prompts of different verifications for test-time inference}
\label{sec:prompt_infer_supp}

\prompt{Chain-of-Thought Verification}{
\{image\} This image is generated by a prompt: \{prompt\}. Please assess the image generation quality step by step. First, breakdown the prompt into multiple visual questions and iteratively answer each question with Yes or No between <think\_start> <think\_end>. Questions should cover all-round details about whether the image accurately represents entity categories, counting of entities, color, spatial relationship in the prompt. Next, output the final result between <answer\_start> <answer\_end>. Output Yes if all multi-choice answers equal yes to show the image has accurate alignment with the prompt. Otherwise answer with No.
}{prompt:cot_v}

\prompt{Outcome Verification}{
\{image\} This image is generated by a prompt: \{prompt\}. Does this image accurately represent the prompt? Please answer yes or no.
}{prompt:out_v}

\prompt{Rule-based Verification}{\{image\} \{question\} Please answer yes or no with detail explanation.}{prompt:rule_v}

\section{Details of Training}
\label{sec:train_detail_supp}

\input{tables/hyper_params}

\subsubsection{Training Parameters}
Details of hyperparameters during each training stage are presented in \Cref{tab:hyperp_train}.

\subsubsection{Training Data Overview}
\label{sec:train_data_supp}
We list the datasets used in our training stages in \Cref{tab:train_data_overview}. Refer to \cref{sec:dpo_data_supp} and \cref{sec:cot_data_supp} for more details about the preference data and \cotshort data.

\subsubsection{Prompts for Generating Pre-Train Data}
\label{sec:train_data_prompt_supp}

We use \cref{prompt:pt_recap} to prompt Qwen2.5VL-7B for generating fine-grained captions for CC-3M, CC-12M, SA-1B and ImageNet that are used in pre-training stages as shown in \Cref{tab:train_data_overview}.

\prompt{Re-caption}{\{image\} What is the content of this image?}{prompt:pt_recap}

\subsubsection{Preference Data Generation for DPO}
\label{sec:dpo_data_supp}

\textbf{PARM.} \textsc{GenEval} metric is used to rate each generated image candidate per prompt. The highest and lowest rated ones are used as the preferred and rejected samples for this prompt.

\textbf{T2I-Comp and SA-1B.}
These prompts are more complex than the prompts of PARM, therefore, it is difficult to rate each generated image using rule-based metrics. We adopt a two-step approach to evaluate the coherence between an image and the prompt. First, we use Qwen2.5-7B to decompose each text prompt into atomic facts represented as questions using \cref{prompt:prompt_2_question}. Then, the image with each decomposed question is fed into Qwen2.5VL-7B with \cref{prompt:prompt_2_ans}. The model responds with \textit{yes} if visual generation passes the fact-check and with \textit{no} otherwise. The final score of an image is calculated by averaging the results of all fact-checks. We take the most and least aligned images per prompt as the preferred and rejected sample pairs.

\prompt{Visual Questions Generation}{Now you need to convert an image description into fine-grained, related visual questions. The questions should comprehensively cover detailed visual facts of entities, attributes (e.g., color, count, texture, shape, and size), and relationships (e.g., spatial and non-spatial) between the entities mentioned in the description. Please complete the task by analyzing each clause in the sentence step by step. For each clause, first raise questions about whether each mentioned entity exists in the image. Then, raise questions about whether the attributes or relationships of the entities are accurately represented in the image. For an image accurately aligned with the description, all questions should be answered with ``yes''; otherwise, they should be answered with ``n''.

Make sure all questions are able to be responded with yes or no and are connected with semicolon. Here are examples:

Example 1:

\quad \textit{description}: three black keys, four chickens and a fabric blanket

\quad \textit{output}: Are there keys?; Are there three keys?; Are the keys black?; Are there chickens?; Are there four chickens?; Is there a blanket?; Is the blanket fabric?

Example 2:

\quad \textit{description}: A person in a blue shirt and red and black apron is using a power tool, likely a drill, to assemble a white cabinet or shelving unit indoors. The floor is covered with light-colored wood or laminate material.

\quad \textit{output}: Is there a person?; Is the person wearing a shirt; Is the shirt blue?; Is the person wearing a apron?; Is the apron red and black?; Is the person using a drill?; Is there a white cabinet or shelving unit?; Is the person using the drill indoors?; Is there light-colored wood on the floor?; Is there laminate material on the floor? 

Example 3:
 
\quad \textit{description}: a large Ferris wheel with a digital clock showing the time as 11:00. The Ferris wheel is located in an urban area, as indicated by the modern buildings in the background. There is also a tree on the left side of the image, partially obscuring the view of the Ferris wheel. The sky appears clear, suggesting a sunny day.
 
\quad \textit{output}: Is there a Ferris wheel?; Is there a digital clock?; Is the digital clock on the Ferris wheel?; Is the digital clock showing the time as 11:00?; Is the Ferris wheel located in an urban area?; Are there modern buildings in the background?; Is there a tree on the left side?; Is the sky clear and sunny?
 
Please convert this image description:\{description\}into fine-grained related visual questions.

}{prompt:prompt_2_question}

\prompt{Visual Fact-Check}{ \{image\} \{question\}
Please answer yes or no without explanation.}{prompt:prompt_2_ans}

\begin{figure*}[ht]
    \centering
    \includegraphics[width=\linewidth]{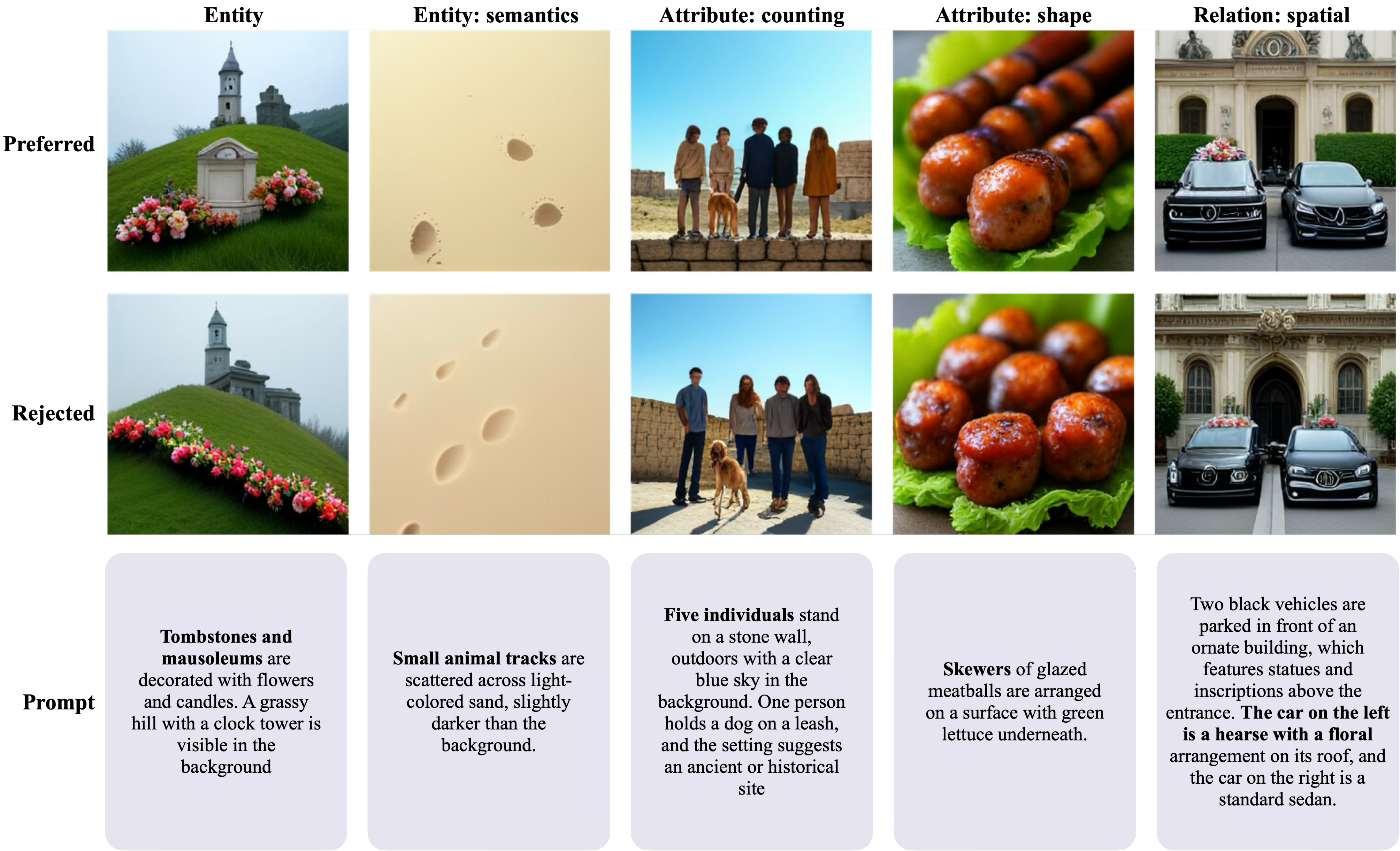}
    \caption{
        \textbf{Visual examples of our generated preference data for DPO training.}
    }
    \label{fig:dpo_example_sa1b}
\end{figure*}

We display some visual results of DPO preference data in   \cref{fig:dpo_example_sa1b}. 

\input{tables/training_data_overview}

\subsubsection{\cotshort Post-Training Data}
\label{sec:cot_data_supp}
We sample 20K preference data from PARM and T2I-Comp in \cref{sec:dpo_data_supp} to construct our \cotshort post-train data and use \cref{prompt:cot_v} to encourage \system to generate CoT reasoning during training. To supervise the training process, we construct the CoT reasoning labels based on decomposed atomic question-answer pairs corresponding to visual facts presented in the image. For PARM, we separate each prompt into fine-grained sub-questions according to the templates originally used for generating the prompt. Rules of \textsc{GenEval} are used to label each sub-question corresponding to the image with \textit{yes} or \textit{no}. For T2I-Comp, we directly use the decomposed question-answers from the preference data. The final answer is \textit{yes} if all the sub-questions are answered with \textit{yes} and it is \textit{no} otherwise. To form the CoT label, the separated question-answers are treated as a thinking process enclosed within special tokens \texttt{<think\_start><think\_end>}, and the final answer resides within \texttt{<answer\_start><answer\_end>}.

\section{More Qualitative Results}
\label{sec:more_qualitative_supp}

\begin{figure}[H]
    \centering
    \includegraphics[width=\linewidth]{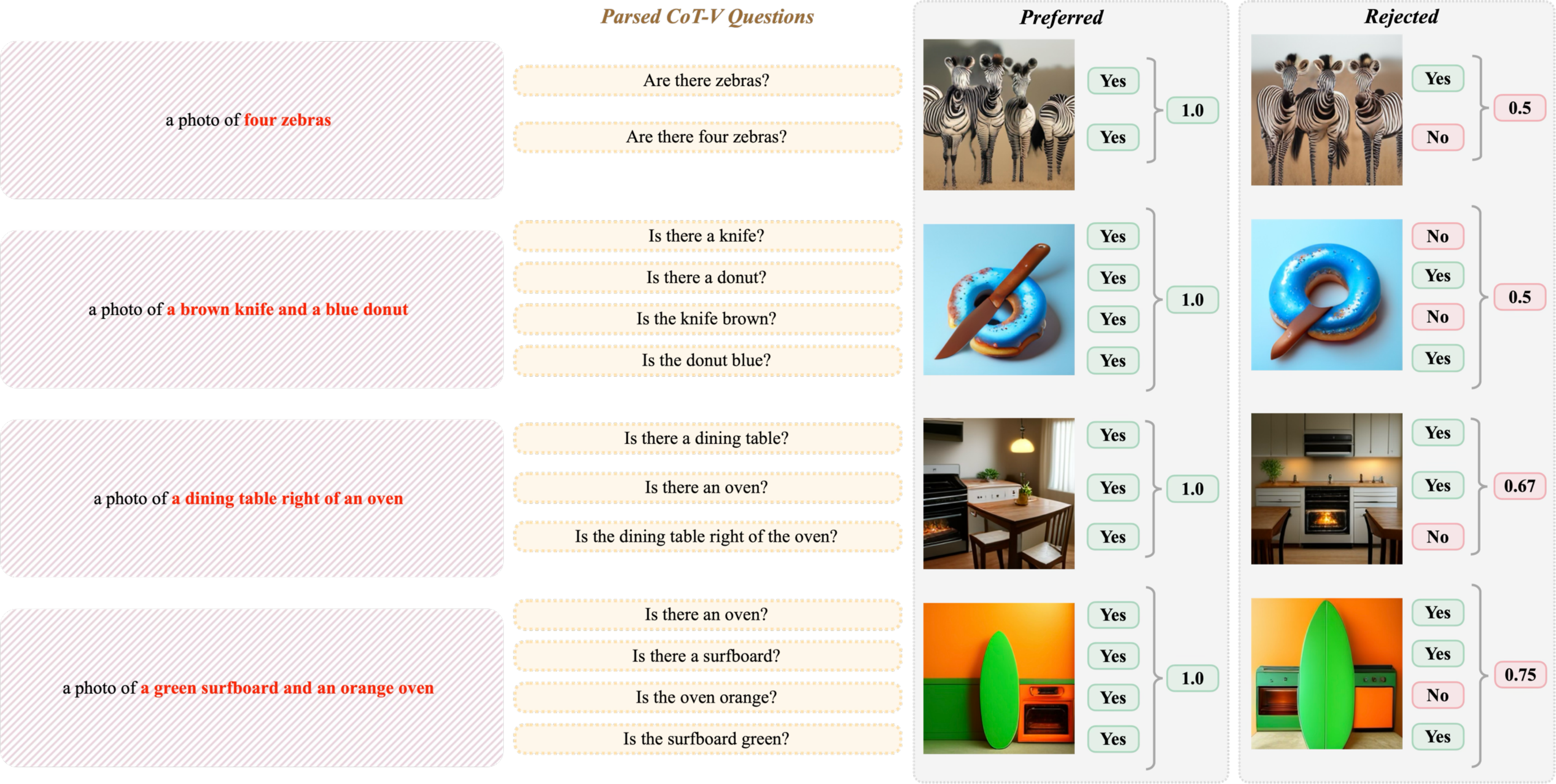}
    \caption{
        \textbf{Successful examples and \cotshort verification on \textsc{Geneval}.}
    }
    \label{fig:cot_v_geneval}
\end{figure}

\begin{figure}[H]
    \centering
    \includegraphics[width=\linewidth]{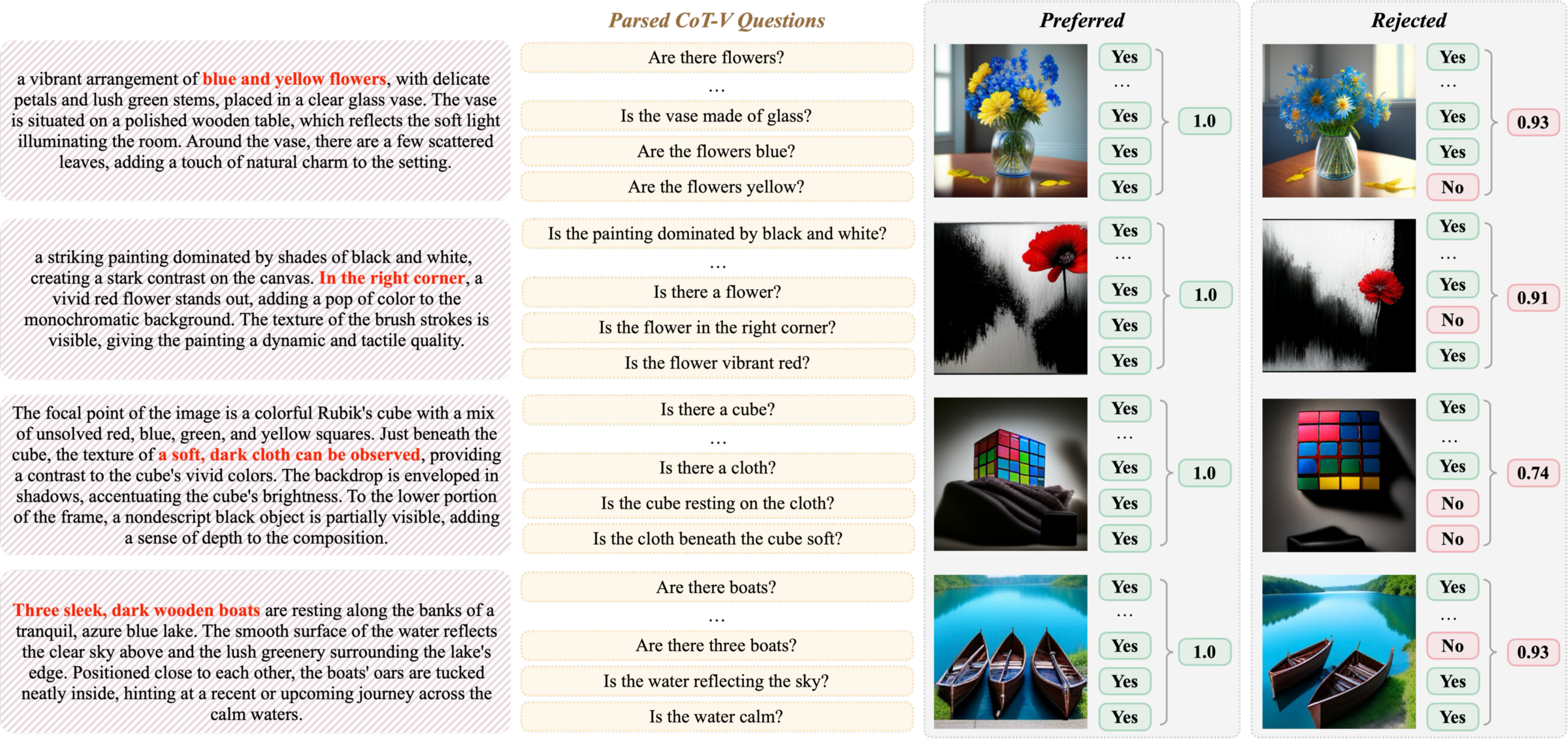}
    \caption{
        \textbf{Successful examples and \cotshort verification on \textsc{DPG-Bench}.}
    }
    \label{fig:cot_v_dpg}
\end{figure}

\begin{figure}[H]
    \centering
    \includegraphics[width=\linewidth]{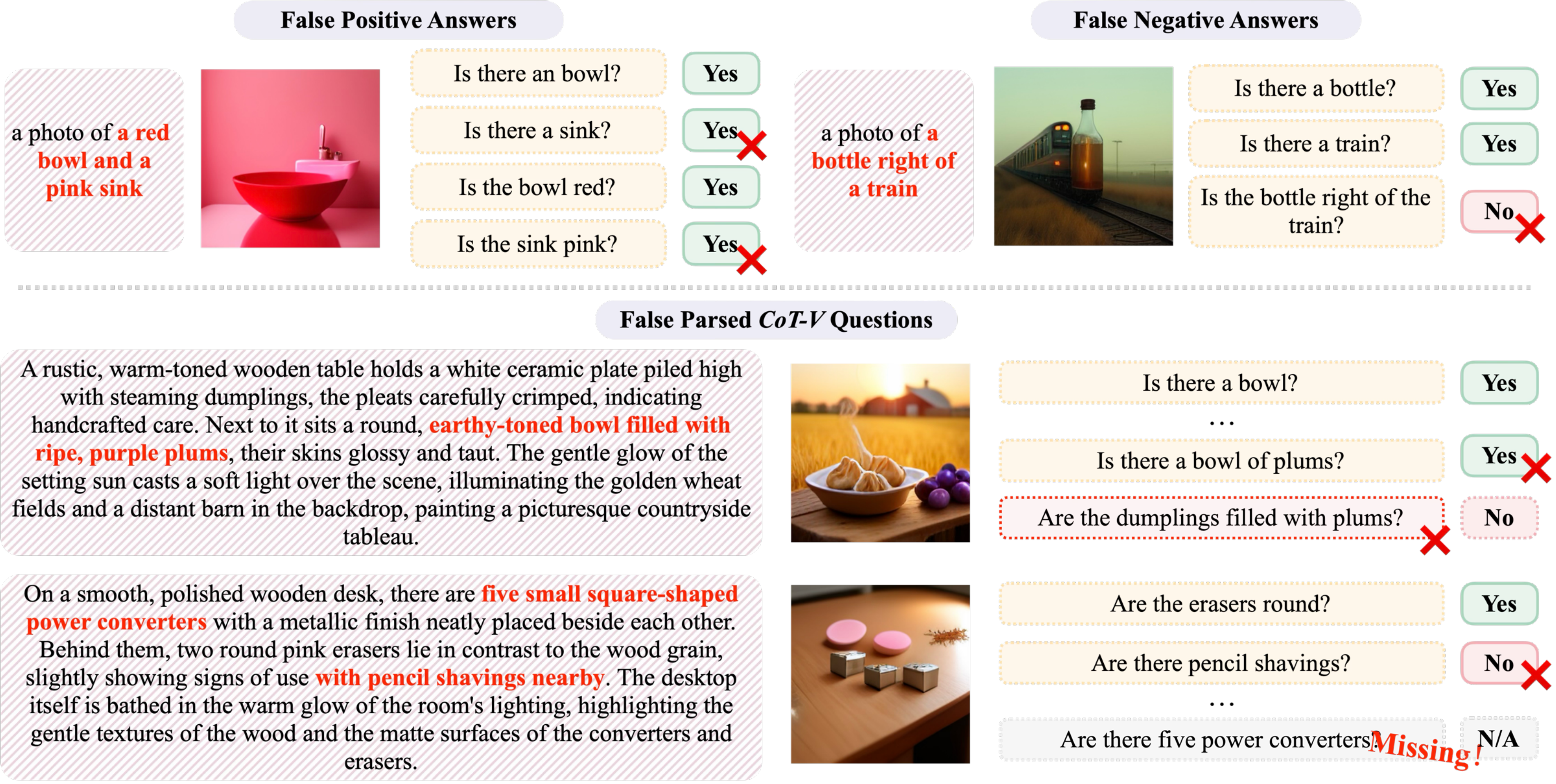}
    \caption{
        \textbf{Failure cases on \textsc{Geneval} and \textsc{DPG-Bench}.} The \textbf{top half} of the image shows failed examples of \cotshort on short prompts.
        The \textbf{bottom half} of the image shows additional cases with bad or missing questions when \cotshort parses the complicated and long prompts.
    }
    \label{fig:cot_v_failure}
\end{figure}

We present qualitative results in \cref{fig:cot_v_geneval} and \cref{fig:cot_v_dpg}. They indicate \cotshort's effectiveness on selecting images that accurately convey the entities, color, counting and spatial relation. However, as failure cases shown in \cref{fig:cot_v_failure}, \cotshort may struggle with hallucination in more difficult cases. Particularly, we acknowledge that \system still falls short of generating an accurate reasoning process given free-form complex prompts. We posit that scaling up the model size or improving our CoT training via reinforcement learning algorithms could improve the capability of reasoning and image generation, and consequently enhance the overall performance of \cotshort.

%% file: tables/sft_ablation.tex
\begin{table}[ht]
\centering
\caption{\textbf{Ablation of SFT stage.} PT-2 Data denotes the text-to-image training data used in the PT-2 Stage. JD and TI denote JourneyDB and text-2-image-2M, respectively. The results are from \systemsft.}
\resizebox{1.0\linewidth}{!}{%
\begin{tabular}{cc|cc|ccccccccc}
\toprule
\textbf{Und Data} & \textbf{Gen Data} & \textbf{GenEval} & \textbf{DPG-Bench}&\textbf{AI2D} & \textbf{GQA} & \textbf{POPE} & \textbf{MMMU} & \textbf{MathVista} & \textbf{RW-QA} & \textbf{ScienceQA} & \textbf{Seedbench} & \textbf{Und Avg.} \\
\midrule
\multirow{2}{*}{SlowFast-LLaVA-1.5} & PT-2 Data &0.56 & 79.67 & 68.3& 	62.4& 	87.5	& 33.3& 42.2& 54.4& 79.6& 70.7 & 62.3
\\
 & JD+TI &0.63 & 82.77& 68.0	&62.5	&87.4	&32.4	&45.2	&58.6	&79.7	&71.1 & 63.1
\\
 \midrule
LLaVA1.5 & JD+TI& 0.64 & 81.82 & 48.7	& 62.8	& 87.4	& 27.1	& 22.1& 	53.7	& 55.5& 	64.0 & 52.7
 \\

\bottomrule
\end{tabular}
}
\label{tab:sft_ablation}
\end{table}

%% file: tables/pretrain_und_data_ablation.tex
\begin{table}[H]
\centering
\caption{\textbf{Impact of the understanding dataset in pre-training stages.} The results are from \systemsft.}
\resizebox{1.0\linewidth}{!}{%
\begin{tabular}{cc|cc|ccccccccc}
\toprule
\makecell[c]{\textbf{Und Data} \\ \textbf{PT-1}} & \makecell[c]{\textbf{Und Data} \\ \textbf{PT-2}} & \textbf{GenEval} & \textbf{DPG-bench}&\textbf{AI2D} & \textbf{GQA} & \textbf{POPE} & \textbf{MMMU} & \textbf{MathVista} & \textbf{RW-QA} & \textbf{ScienceQA} & \textbf{Seedbench} & \textbf{Und Avg.} \\
\midrule
\ding{55} & \ding{55} & 0.61 & 82.51 &60.5 &59.6 &87.4 &30.9 &38.1 &49.0 &72.0 & 66.1 & 58.0\\
\ding{51} & \ding{51} & 0.63 &82.75 &68.0 &62.5 &87.4 &32.4 &45.2 &58.6 &79.7 & 71.1 & 63.1\\
\bottomrule
\end{tabular}
}
\label{tab:pretrain_und_data_ablation}
\end{table}

%% file: tables/pretrain1_ablation.tex
\begin{table}[H]
\centering
\caption{\textbf{Ablation of PT-1 stage.} Cls and Recap indicate class names and high-quality captions are used for generating images, respectively. The reported numbers of the understanding tasks are from \systemsft.}
\resizebox{1.0\linewidth}{!}{%
\begin{tabular}{lc|cc|ccccccccc}
\toprule
\textbf{Stage I}  & \textbf{Gen Data}  & \textbf{GenEval}  & \textbf{DPG-bench} &\textbf{AI2D}  & \textbf{GQA}  & \textbf{POPE}  & \textbf{MMMU}  & \textbf{MathVista}  & \textbf{RW-QA}  & \textbf{ScienceQA}  & \textbf{Seedbench}  &\textbf{Und Avg.} \\
\midrule
\ding{55} & --& 0.64	&82.26	&70.3	&62.5	&87.9	&33.7	&45.7	&54.2	&80.5	&71.7 & 63.3
\\
\ding{51} & ImageNet (Cls)&0.63&	82.75	&67.0	&62.5	&88.0	&31.4	&41.4	&53.6	&79.8	&71.1 &61.8

\\
\ding{51} & ImageNet (Recap) &0.63	&82.75	&68.0	&62.5	&87.4	&32.4	&45.2	&58.6	&79.7	&71.1 & 63.1\\

\bottomrule
\end{tabular}
}
\label{tab:pretrain1_ablation}
\end{table}

%% file: tables/pretrain2_ablation.tex
\begin{table}[H]
\centering
\caption{\textbf{Ablation of PT-2 stage.} The reported results are from \systemsft.}
\resizebox{1.0\linewidth}{!}{%
\begin{tabular}{cc|cc|ccccccccc}
\toprule
\textbf{Stage II}  & \textbf{Gen \& Und Data}  & \textbf{GenEval}  & \textbf{DPG-bench} &\textbf{AI2D}  & \textbf{GQA}  & \textbf{POPE}  & \textbf{MMMU}  & \textbf{MathVista}  & \textbf{RW-QA}  & \textbf{ScienceQA}  & \textbf{Seedbench}  &\textbf{Und Avg.} \\
\midrule
\ding{55} & -- & 0.58&	79.25	&70.2	&61.9	&87.3	&31.6	&47.0	&54.9	&82.5	&71.2 & 63.3\\
\ding{51} & CC+SA+IMN &0.59	&80.64	&64.5	&61.6	&87.9	&30.8	&40.9	&54.2	&77.0	&69.6 & 60.8
\\
\ding{51} & (SA+IMN) (Recap) &0.64	&82.63	&67.2	&62.4	&87.5	&31.0	&40.8	&56.5	&79.7	&71.1 & 62.0
\\
\ding{51} & (CC+IMN) (Recap) &0.63	&82.75	&67.9	&62.1	&87.8	&31.6	&42.0	&58.2	&80.3	&70.8 & 62.6
\\
\ding{51} & (CC+SA) (Recap) & 0.62 & 82.34	&68.6	&62.2	&87.4	&30.6	&45.4	&59.3	&80.5	&70.8 & 63.1
\\
\ding{51} & (CC+SA+IMN) (Recap) &0.63	&82.75	&68.0	&62.5	&87.4	&32.4	&45.2	&58.6	&79.7	&71.1 & 63.1
\\
\bottomrule
\end{tabular}
}
\label{tab:pretrain2_ablation}
\end{table}

%% file: tables/geneval_final.tex
\begingroup
    \begin{table*}[ht]
    \caption{
        \textbf{Comparison with state-of-the-art models on the \textbf{GenEval} benchmark.}
    }
    \centering
    \footnotesize
    \addtolength{\tabcolsep}{-3pt}
    \ra{1}
    \resizebox{\linewidth}{!}{%
    \begin{tabular}{lc|ccccccc}
    \toprule
    \textbf{Model} & \textbf{\#Params} & \textbf{Single Obj.}& \textbf{Two Obj.}  & \textbf{Counting}& \textbf{Colors} & \textbf{Position} & \textbf{Color Attri.} & \textbf{Overall$\uparrow$}\\
    \cmidrule{1-2}\cmidrule{3-9}
    \multicolumn{9}{c}{\textit{Text-to-Image Generation Models}}\\
    \midrule
    DALLE-2~\cite{ramesh2022hierarchical} & 6.5B & 0.94 & 0.66 & 0.49 & 0.77 & 0.10 & 0.19 & 0.52 \\
    DALLE-3\cite{betker2023improving} & - & 0.96 & 0.87 & 0.47 & 0.83 & 0.43 & 0.45 & 0.67 \\
    Emu3~\cite{wang2024emu3} & 8B & 0.98 & 0.71 & 0.34 & 0.81 & 0.17 & 0.21 & 0.54 \\ 
    SDXL~\cite{podellsdxl} & 2.6B &0.98 &0.74 &0.39 &0.85 &0.15 &0.23 &0.55 \\
    SimpleAR~\cite{wang2025simplear} & 1.5B &- &0.90  &- &-&0.28 &0.45 &0.63 \\
    Infinity~\cite{han2024infinity} & 2B & - & 0.85 & -& - & 0.49 & 0.57 & 0.73 \\
    \midrule 
    \multicolumn{9}{c}{\textit{Unified MLLMs}}\\
    \midrule
Show-o~\cite{xie2024show}& 1.3B &  0.95 & 0.52 & 0.49 & 0.82 & 0.11 & 0.28 & 0.53 \\
 Janus~\cite{wu2024janus} & 1.3B& 0.97 & 0.68 & 0.30 & 0.84 & 0.46 & 0.42 & 0.61 \\
 Janus-Pro~\cite{chen2025janus} & 1.5B&  0.98 & 0.82 & 0.51 & 0.89 & 0.65 & 0.56 & 0.73 \\
 ILLUME~\cite{wang2024illume} & 7B  & 0.99 & 0.86 & 0.45 & 0.71 & 0.39 & 0.28 & 0.61 \\
 UniToken~\cite{jiao2025unitoken} & 7B & 0.99 & 0.80  &0.35 & 0.84  &0.38  &0.39 & 0.63 \\
 VARGPT-v1.1~\cite{zhuang2025vargpt} & 9B & 0.96 &0.53 &0.48 &0.83 &0.13 &0.21 &0.53\\
 TokenFlow-XL~\cite{qu2024tokenflow} & 13B &  0.93 & 0.72 & 0.45 & 0.82 & 0.45 & 0.42 & 0.63 \\
 \rowcolor{lightblue} \system  & 1.5B & 1.00 & 0.94& 0.78 & 0.87& 0.57 &0.54 &\textbf{0.78}  \\
%    \midrule
%Janus-Pro~\cite{chen2025janus} & 7B& 0.99 &0.89 &0.59 &0.90 &0.79 &0.66 &0.80 \\   
    \bottomrule
    \end{tabular}
    }
    \label{tab:geneval_cmp}
\end{table*}
\endgroup

%% file: tables/dpg_final.tex
\begingroup
    \begin{table*}[ht]
    \caption{
        \textbf{Comparison with state-of-the-art models on the \textbf{DPG-bench} benchmark}.
    }
    \centering
    \footnotesize
    \addtolength{\tabcolsep}{1pt}
    \ra{1}
    \resizebox{\linewidth}{!}{%
    \begin{tabular}{lc|ccccccc}
    \toprule
    \textbf{Model} & \textbf{\#Params} & \textbf{Global}& \textbf{Entity}  & \textbf{Attribute}& \textbf{Relation} & \textbf{Other} & \textbf{Overall$\uparrow$}\\
    \cmidrule{1-2}\cmidrule{3-8}
    \multicolumn{8}{c}{\textit{Text-to-Image Generation Models}}\\
    \midrule    
    Hunyuan-DiT~\cite{li2024hunyuan}& - & 84.59& 80.59& 88.01 &74.36& 86.41 &78.87\\
    DALLE-3\cite{betker2023improving} & - & 90.97 &89.61 &88.39 &90.58 &89.83 &83.50\\
    Emu3~\cite{wang2024emu3} & 8B &
    85.21 &86.68 &86.84 &90.22& 83.15 &80.60\\
    SDXL~\cite{podellsdxl} & 2.6B &83.27 & 82.43 & 80.91 & 86.76 & 80.41 & 74.65 \\
    SimpleAR~\cite{wang2025simplear} & 1.5B & 87.97 &- &- & 88.33 &- & 81.97 \\
    Infinity~\cite{han2024infinity}& 2B & 93.11 & - & - & 90.76 & - &  83.46\\
    \midrule
    \multicolumn{8}{c}{\textit{Unified MLLMs}}\\
    \midrule
    Show-o*~\cite{xie2024show}& 1.3B & 80.39 & 80.94 &82.17 & 83.36 & 82.88& 71.70 \\
    Janus~\cite{wu2024janus} & 1.3B& 82.33 &87.38 &87.70 &85.46 &86.41 &79.68\\
    Janus-Pro~\cite{chen2025janus} & 1.5B& 87.58 &88.63& 88.17 &88.98 &88.30 &82.63\\
    VARGPT-v1.1~\cite{zhuang2025vargpt} & 9B & 84.83 & 82.80&  84.95 &88.13 & 87.70&  78.59\\
    TokenFlow-XL~\cite{qu2024tokenflow} & 13B & 78.72 & 79.22& 81.29& 85.22& 71.20 & 73.38\\
    %\rowcolor{lightblue} \system & 1.5B & 91.53 & 90.39 &90.30 & 91.09 &90.86 & \textbf{84.89}\\
    \rowcolor{lightblue} \system & 1.5B& 91.95 & 89.68 & 90.90 & 92.04 & 90.91 & \textbf{85.19}\\ 
    \midrule
    \end{tabular}
    }
    \label{tab:dpg_cmp}
\end{table*}
\endgroup

%% file: tables/hyper_params.tex
\begingroup
\begin{table*}[ht]
    \caption{
        \textbf{Hyperparameter setup for different training stages of \system.} Data ratio refers to the ratio of image understanding data, pure text data, and image generation data.
    }
    \ra{1.2}
    \centering
    \footnotesize
    \addtolength{\tabcolsep}{-1pt} 
    % \resizebox{\linewidth}{!}{%
    \begin{tabular}{l|ccc|c|c}
    \toprule
    \textbf{Hyperparameters} & \textbf{PT-1} & \textbf{PT-2} & \textbf{SFT} & \textbf{DPO} & \textbf{\cotshort Post-Training} \\
    \hline
    Learning rate         & $1.0\times10^{-4}$   & $1.0\times10^{-4}$ & $1.0\times10^{-3}$   & $1.0\times10^{-5}$ & $1.0\times10^{-5}$ \\
    LR scheduler       & Cosine   & Cosine & Cosine  & Cosine & Cosine\\
    Weight decay      & 0.01   &0.01 & 0.05 & 0.05 & 0.05  \\
    Gradient clip       & 1.0 & 1.0 & 1.0 & 1.0 & 1.0 \\
    Optimizer & AdamW  & AdamW & AdamW & AdamW & AdamW\\
    Warm-up steps  & 6000 & 5000 & 1000 & 500 & 0  \\
    Training steps        & 150k & 400k & 146k & 1.6k &0.5k  \\
    H100 hours & 1.0k & 2.8k & 240 & 5 & 0.7\\
    Batch size        & 896 & 512 & 64 & 80 & 64\\
    Data ratio        & 2:1:4 & 2:1:4 & 4:1:3 & -:-:1 & 1:-:-  \\
    \midrule
    \end{tabular}
    % }
\label{tab:hyperp_train}
\vspace{-0.2in}
\end{table*}
\endgroup

%% file: tables/training_data_overview.tex
\begin{table}[ht]
\centering
\footnotesize
\caption{Training data overview. CC, SA, IMN, JD, T2I indicate CC-3M\&CC-12M, SA-1B, ImageNet, JourneyDB, text-2-image-2M, respectively. Recap denotes that the images are re-captioned using Qwen2.5VL-7B.}
% \resizebox{0.9\linewidth}{!}{%
\begin{tabular}{l|ccc}
\toprule
Stage & Gen Data & Und Data & Text-only\\
\midrule
PT-1 & IMN (Recap) & (CC+SA+IMN) (Recap) & RefinedWeb\\
PT-2 & (CC+SA+IMN) (Recap) & (CC+SA+IMN) (Recap) & RefinedWeb\\
SFT & JD+T2I & SF-LLaVA1.5 (Image Mixture) \cite{xu2025slowfast} & RefinedWeb\\
DPO & Preference Data & -- & -- \\
\cotshort & -- & \cotshort data & --\\
\bottomrule
\end{tabular}
% }
\label{tab:train_data_overview}
\end{table}